\pdfoutput=1

\documentclass[11pt]{article}

\usepackage[preprint]{acl}

\usepackage{times}
\usepackage{latexsym}
\usepackage{amsmath}
\usepackage{multirow}
\usepackage{tabularx}
\usepackage{xcolor}

\usepackage[T1]{fontenc}

\usepackage[utf8]{inputenc}

\usepackage{microtype}

\usepackage{inconsolata}

\usepackage{graphicx}

\title{Ensembling Large Language Models with Process Reward-Guided Tree Search for Better Complex Reasoning}

\author{
  Sungjin Park\thanks{Work done during an internship at Microsoft Research Asia.}$^{1,2}$, Xiao Liu$^{2}$, Yeyun Gong$^{2}$, Edward Choi$^{1}$ \\
  $^{1}$KAIST AI \\
  $^{2}$Microsoft Research \\
  \texttt{\{zxznm, edwardchoi\}@kaist.ac.kr, \{xiao.liu.msrasia, yegong\}@microsoft.com}\\
}

\begin{document}
\maketitle
\begin{abstract}
Despite recent advances in large language models, open-source models often struggle to consistently perform well on complex reasoning tasks. 
Existing ensemble methods, whether applied at the token or output levels, fail to address these challenges.
In response, we present Language model Ensemble with Monte Carlo Tree Search (LE-MCTS), a novel framework for process-level ensembling of language models. 
LE-MCTS formulates step-by-step reasoning with an ensemble of language models as a Markov decision process. 
In this framework, states represent intermediate reasoning paths, while actions consist of generating the next reasoning step using one of the language models selected from a predefined pool. 
Guided by a process-based reward model, LE-MCTS performs a tree search over the reasoning steps generated by different language models, identifying the most accurate reasoning chain. 
Experimental results on five mathematical reasoning benchmarks demonstrate that our approach outperforms both single language model decoding algorithms and language model ensemble methods.
Notably, LE-MCTS improves performance by 3.6\% and 4.3\% on the MATH and MQA datasets, respectively, highlighting its effectiveness in solving complex reasoning problems.
\end{abstract}

\section{Introduction}
\label{sec:intro}
Large language models (LLMs) have demonstrated superior performance across a range of tasks, primarily due to their large capacity and high-quality training data. 
However, unlike prominent closed-source LLMs such as GPT-4~\citep{openai-etal-2023-gpt4} and Gemini-1.5~\cite{gemini-etal-2024-gemini}, open-source models like Mistral~\citep{jiang-etal-2023-mistral}, LLaMA-3~\citep{dubey-etal-2024-llama3}, and Gemma-2~\citep{gemma-etal-2024-gemma2} are constrained by factors such as data availability, architecture, and hyperparameter choices.
As a result, they exhibit different strengths and weaknesses~\citep{jiang-etal-2023-llm}. 

\begin{figure}[t]
    \centering
    \includegraphics[width=\linewidth]{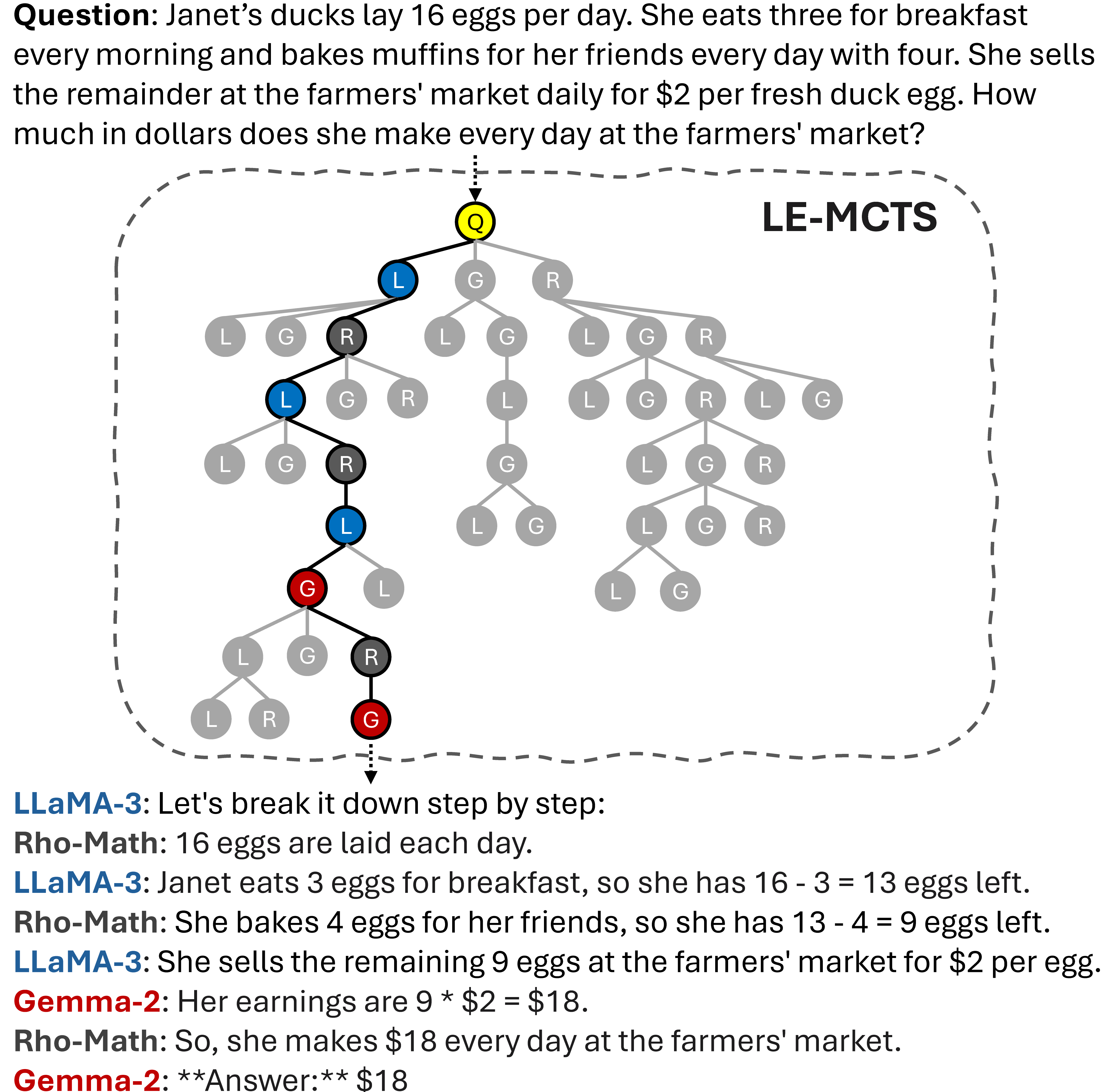}
    \caption{\textbf{Example output of LE-MCTS.} The reasoning steps in the LE-MCTS output can be generated by different LLMs. We highlight the root node in yellow and apply the same color coding to the corresponding nodes and the language model.}
    \label{fig:method:lemcts}
\end{figure}

Language model ensembling is a well-known approach for creating a versatile model by combining weaker language models. 
Previous studies on LM ensembling have focused on merging language models at the token and output levels.
Token-level approaches have merged the output logits or probabilities of language models, using perplexity-based weighted averaging~\citep{liu-etal-2024-coolfusion, mavromatis-etal-2024-packllm} and vocabulary projection~\citep{xu-etal-2024-bridging}.
Although token-level ensemble methods offer fine-grained fusion of language models, they face several constraints related to tokenizer vocabularies and model architectures, requiring the training of additional projection matrices~\citep{xu-etal-2024-bridging} to mitigate these.
Output-level approaches have involved ensembling fully generated outputs, either by ranking multiple outputs to select the best one~\citep{farinhas-etal-2023-empirical, jiang-etal-2023-llm} or by using an additional fusion model to fuse them~\citep{jiang-etal-2023-llm}.
Although output-level ensembling can be applied to any language model, it cannot produce a correct answer if all candidate outputs are flawed~\citep{xu-etal-2024-bridging}.

Our experiments in Table~\ref{tab:main_result} show that LLMs have varying levels of expertise across different types of math problems. 
Such disparities cannot be easily mitigated by applying decoding algorithms designed for a single LLM.
Furthermore, the results show that both the existing token-level and output-level LM ensemble methods perform worse than a single language model. 
These results clearly demonstrate that current LM ensemble methods are particularly ineffective in solving complex reasoning problems, pointing to the need for a framework specifically designed to handle such tasks and to ensure consistently high performance across diverse reasoning problems.

To address the limitations of token- and output-level ensembling, we propose a process-level language model ensemble framework tailored to complex reasoning tasks. 
Many complex reasoning problems can be solved step by step, and evaluating each reasoning step individually allows the model to correct errors early, thus guiding the decoding process toward more accurate solutions~\cite{yao-etal-2024-tree, besta-etal-2024-graph, yu-etal-2024-ovm, ma-etal-2023-let}. 
This approach is further supported by recent advances in process-based reward models (PRMs)~\citep{lightman-etal-2023-prm800k, wang-etal-2024-math, lu-etal-2024-autocv}, which make it possible to measure the correctness of each intermediate reasoning step as a scalar value.
Based on the advantages of process-level reasoning, process-level ensembling offers finer and more efficient control over generation compared to output-level ensembling, as it allows for the correction of intermediate reasoning steps without regenerating entire solutions. 
Additionally, process-level ensembling is less constrained than token-level ensemble approaches, as it eliminates the need to match vocabulary and architecture.

We present Language model Ensemble with Monte Carlo Tree Search (LE-MCTS), a pioneering framework for process-level ensembling of language models. 
We formulate the step-by-step reasoning involving multiple LLMs as a Markov decision process (MDP)~\citep{bellman-etal-1957-mdp}. 
Specifically, the state is defined as the intermediate reasoning steps, and the action is defined as generating the next reasoning step using one of the language models selected from a predefined pool of LLMs.
Our MCTS algorithm, inspired by AlphaZero~\citep{silver-etal-2020-alphazero}, performs a tree search over the unified space of reasoning steps generated by different LLMs. 
By following the guidance from PRM, we can obtain the reasoning chain that is likely to be the most accurate among the possible combinations of reasoning steps generated by the LLMs.

We evaluate LE-MCTS on five math reasoning benchmarks: GSM8K~\citep{cobbe-etal-2021-gsm8k}, MATH~\citep{hendrycks-etal-2021-math}, SVAMP~\cite{patel-etal-2021-nlp}, ASDiv~\cite{miao-etal-2020-diverse}, and MQA~\cite{amini-etal-2019-mqa}. 
Our results show that LE-MCTS consistently outperforms or matches existing LM ensemble approaches across all tasks. 
We also evaluate a straightforward integration of process reward-guided decoding algorithms with the LM ensemble. 
Although these models perform well on grade school math problems, they fail on more complex datasets, such as MATH and MQA. 
In contrast, LE-MCTS improves performance by 3.6\% and 4.3\% on MATH and MQA, respectively, compared to the second-best models.
These experimental results highlight the effectiveness of LE-MCTS in tackling complex reasoning problems.
\section{Methods}
\label{sec:method}
\begin{figure*}[t]
    \centering
    \includegraphics[width=\textwidth]{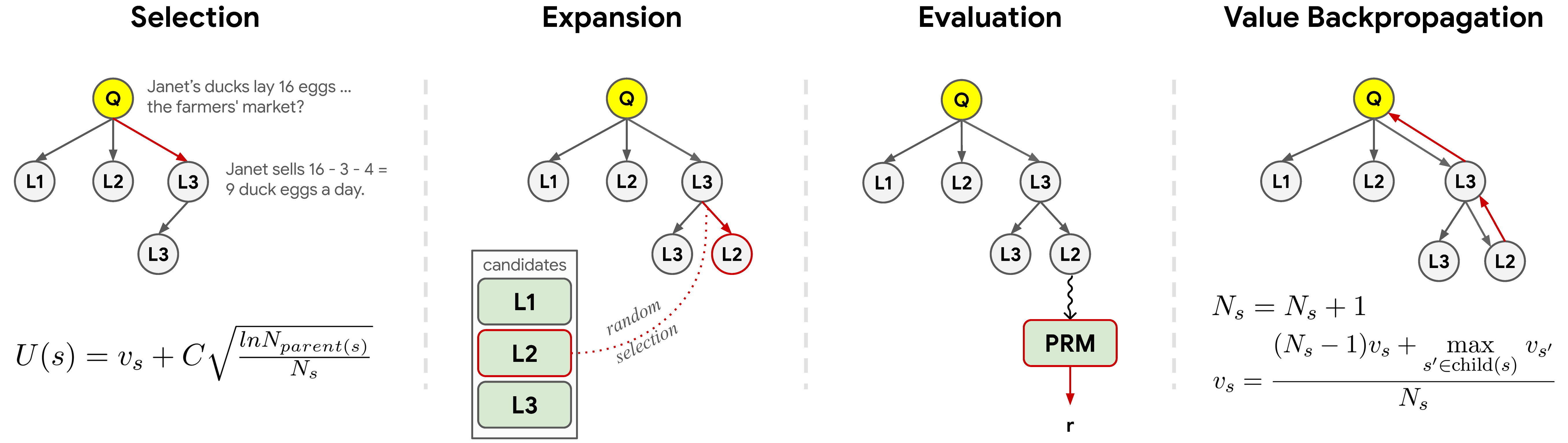}
    \caption{
    \textbf{Single iteration of LE-MCTS.} This example illustrates an ensemble of three LLMs. 
    The iteration is repeated until the maximum number of iterations, \(n_{iter}\), is reached or no further nodes in the tree can be expanded.
    }
    \label{fig:method:lemcts}
\end{figure*}
First, we introduce the problem formulation and key notation. We then describe our LE-MCTS algorithm. 

\subsection{Problem Setup}
\label{text:method:setup}

Given an input problem \( q \) and \( L \) language models \( \{\pi_1, \dots, \pi_L\} \), a language model \( \pi_l \) can generate a complete output \( o \) or a reasoning step \( p_k \sim \pi_l(\cdot \mid q, p_{1:k-1}) \), where \( p_{1:k-1} \) is the sequence of previously generated reasoning steps up to \( p_k \). 
We define each sentence ending with a newline character as a reasoning step.
This setup allows us to compare reasoning steps from different language models, \( \{p^1_k, \dots, p^L_k\} \), which are generated from the same intermediate reasoning trajectory \( p_{1:k-1} \). 
Our objective is to find the optimal combination of reasoning steps generated by different LMs, denoted as \( o^* = p^*_{1:K} \), where \( p_K \) is the final reasoning step that includes the answer. 
Ideally, we can obtain \( o^* \) by evaluating all elements in the set \( P = \prod^{K}_{k=1} P_k \), where \( P_k = \{p^1_k, \dots, p^L_k\} \).
While \( o^* \) offers a better solution compared to a single LM output \( o^l \), it becomes infeasible to evaluate the entire \( P \) as \( L \) and \( K \) increase. 

To address this issue, we model the step-by-step reasoning problem as a Markov decision process (MDP)~\citep{bellman-etal-1957-mdp} and solve it using Monte Carlo Tree Search (MCTS)~\citep{coulom-etal-2006-mcts}. 
Specifically, the root node represents the input \( q \), while each child node \( s \) corresponds to an intermediate reasoning trajectory \( p_{1:k} \), storing its value \( v_s \) and visit count \( N_s \).
The value \( v_s \) represents the accuracy of the intermediate reasoning trajectory.
An action is defined as generating \( p^l_k \) from \( p_{1:k-1} \) using \( \pi_l \), selected from \(\{\pi_1,...,\pi_L\}\). 
PRM computes the reward to guide MCTS towards maximizing the process reward for both intermediate reasoning steps and the final output.
In the following sections, we introduce how LE-MCTS effectively integrates reasoning steps from various LMs into the tree and performs a lookahead search in this setup.

\subsection{LE-MCTS}
\label{text:method:lemcts}
Each iteration of our LE-MCTS algorithm consists of four stages: selection, expansion, evaluation, and value backpropagation. 
LE-MCTS repeats the iterations until the number of iterations reaches the maximum value \(n_{iter}\), or no further nodes in the tree can be expanded. 
A pseudocode for LE-MCTS is provided in Table~\ref{tab:pseudo_code}.
\medskip
\par \noindent \textbf{Selection} \;
LE-MCTS begins each iteration with the selection phase, starting at the root node and hierarchically selecting child nodes until a leaf node \( s_\text{leaf} \) is reached. To incorporate the quality values of nodes, we use the UCT algorithm~\citep{kocsis-etal-2006-uct} for selecting a child node:
\begin{equation}
    U(s) = v_s + C\sqrt{\frac{\ln N_{\text{parent}(s)}}{N_s}}
\end{equation}
where \( N_{\text{parent}(s)} \) is the visit count of \( s \)'s parent node, and \( C \) is a constant controlling the exploration-exploitation trade-off. 
At each intermediate node, we select the child with the highest \( U(s) \). 
A higher \( C \) promotes exploration of underexplored nodes, while a lower \( C \) favors searching for high-value nodes first. 
Terminal state nodes are excluded from selection, as the goal of LE-MCTS selection is to identify the best incomplete intermediate reasoning trajectory.
The selection enables LE-MCTS to explore high-accuracy trajectories effectively without exhaustively examining the entire set \( P \).
\medskip
\par \noindent \textbf{Expansion} \;
After selecting the leaf node \( s_{\text{leaf}} \), we add a new child node to it. 
For simplicity, we assume \( s_{\text{leaf}} \) corresponds to the intermediate reasoning trajectory \( p_{1:k-1} \) in the following sections.

First, we randomly select a language model \( \pi_l \) from the pool \( \{\pi_1, \dots, \pi_L\} \). 
We then greedily decode the next reasoning step \( p_k \) using \( \pi_l \) until a newline character appears:
\begin{equation}
    p_{k,t} = \underset{w \in V}{\arg\max} \; \pi_l(w \mid p_{k,<t}; q, p_{1:k-1})
\end{equation}
where \( V \) is the vocabulary.
The expansion phase plays a key role in the process-level ensemble because it introduces new reasoning steps generated by different language models into the tree.
We do not expand fully-expanded nodes that meet any of the following criteria:
\begin{itemize}
    \item The number of child nodes reaches the constant \( n_{\text{child}} \)
    \item \( v_s - \underset{s' \in \text{child}(s)}{\max} \; v_{s'} < \varepsilon \)
\end{itemize}
The second criterion encourages LE-MCTS to explore deeper reasoning paths by prioritizing depth over breadth in its expansions.
In complex reasoning problems, the length of the reasoning chain tends to increase with difficulty of the problem.
Therefore, this criterion enables LE-MCTS to explore sufficiently deep reasoning trajectories within a limited number of MCTS iterations.

\medskip
\par \noindent \textbf{Evaluation} \;
In the evaluation phase, we compute the reward of the reasoning step \( p_{k} \). 
Following the approach used in AlphaZero and its variants~\citep{silver-etal-2020-alphazero, feng-etal-2023-alphazerollm}, we do not perform any rollouts. 
Instead, we directly employ the process-based reward model (PRM) to compute the reward. 
Specifically, we utilize a pre-trained PRM \(\phi\) from Math Shepherd~\citep{wang-etal-2024-math}, which takes the reasoning step \( p_{k} \) and the problem \( q \) as inputs, predicting the process reward \(r_k=\phi(q, p_k)\).

One advantage of performing rollouts in our setup is that it enables the use of an outcome-based reward model (ORM) to compute the reward.
However, as demonstrated by \citet{uesato-etal-2022-oprm}, both PRM and ORM emulate process-based feedback and achieve similar performance. 
Therefore, we decide not to perform rollouts, as the performance improvement with ORM is minimal, while the execution time increases by approximately five to tenfold.

\medskip
\par \noindent \textbf{Value Backpropagation} \;
After expanding and evaluating a leaf node, the statistics are propagated up the tree, and each node visited during the selection phase updates its value and visit count. 
The standard practice~\citep{browne-etal-2012-survey} for value backpropagation is increasing the visit count \(N_s\) by 1 and then updating the node's value according to the following equation:
\begin{equation}
    v_s = \frac{(N_s-1)v_s + r_k}{N_s}
\end{equation}
With the standard backpropagation strategy, the reasoning steps generated by different language models contribute equally to \( v_s \).
However, in our setup, it is sufficient to identify at least one intermediate reasoning trajectory where any language model can generate a viable subsequent reasoning path.
Thus, a node is considered acceptable as long as it has at least one child with a high value, even if the others have low values.

To address this, we propose a new backpropagation strategy, \textit{optimistic backpropagation}, which updates the node's value based on the maximum value among its child nodes instead of the reward:
\begin{equation}
    v_s = \frac{(N_s-1)v_s + \underset{s' \in \text{child}(s)}{\max}\; v_{s'}}{N_s}
\end{equation}
This optimistic backpropagation strategy disregards the signals from low-valued sibling nodes and directs LE-MCTS to focus on the most promising reasoning path.
 
\begin{table}[t]
\centering
\resizebox{\columnwidth}{!}{%
\begin{tabular}{l}
\hline
\textbf{Algorithm: }\textbf{LE-MCTS} \\
\hline
\textbf{Input}: input $q$, language models $\{\pi_1, \dots, \pi_L\}$, max MCTS iterations $n_{iter}$, \\
\ \,\quad UCT constant $C$, max \# child nodes $n_{child}$, threshold $\varepsilon$, PRM $\phi$ \\
\ \texttt{// Initialize} \\
\ 1: \(s_0\) \(\leftarrow\) \text{CreateNode}(\(T,q\)) \\
\ 2: \textbf{for} $i = 1, ..., n_{iter}$ \textbf{do}\\
\ 3: \quad $s \leftarrow s_0$ \\
\ \quad\quad\! \texttt{// Selection} \\ 
\ 4: \quad \textbf{while} $s$ is not a leaf node \textbf{do} \\
\ 5: \quad\quad \( S \leftarrow \{\} \) \\ 
\ 6: \quad\quad \textbf{for} \( s' \in \text{child}(s) \) \textbf{do} \\
\ 7: \quad\quad\quad \textbf{if} $n(\text{child}(s')) < n_{child}$ \textbf{and} $v_s' - \max\limits_{s'' \in \text{child}(s')} v_{s''} \geq \varepsilon$ \textbf{then} \\
\ 8: \quad\quad\quad\quad \( S \leftarrow S+\{s'\} \) \\ 
\ 9: \quad\quad\quad \textbf{end if} \\
\ 10: \quad\quad \textbf{end for} \\
\ 11: \quad\quad $s \leftarrow \underset{s' \in S}{\arg\max}\; v_{s'} + C \sqrt{\dfrac{\ln N_s}{N_{s'}}}$\\
\ 12: \quad \textbf{end while}\\
\ \quad\quad\! \texttt{// Expansion} \\ 
\ 13: \quad \( \pi_l \leftarrow \text{RandomSelect}(\{\pi_1,...,\pi_L\})\) \\
\ 14: \quad \( p_{1:k-1} \leftarrow \text{GetPath}(s) \)  \\
\ 15: \quad \textbf{while} $p_{k,t}$ is not \texttt{\textbackslash n} \textbf{do}\\
\ 16: \quad\quad $p_{k,t} \leftarrow \underset{w \in V}{\arg\max}\; \pi_l(w \mid p_{k,<t}; q, p_{1:k-1})$ \\
\ 17: \quad \textbf{end while} \\
\ 18: \quad $ s' \leftarrow \text{CreateNode}(T,\{p_{1:k-1},p_k\})$ \\
\ \quad\quad\; \texttt{// Evaluation and Value Backpropagation} \\ 
\ 19: \quad \( v_{s'} \leftarrow \phi(q, p_k), N_{s'} \leftarrow 1\) \\ 
\ 20: \quad \textbf{while} \( s' \) is not a root node \textbf{do} \\
\ 21: \quad\quad \( s' \leftarrow \text{GetParent}(s') \) \\
\ 22: \quad\quad \( N_{s'} \leftarrow N_{s'}+1\) \\
\ 23: \quad\quad $v_{s'} \leftarrow \dfrac{(N_{s'} - 1) v_{s'} + \underset{s'' \in \text{child}(s')}{\max}\; v_{s''}}{N_{s'}}$ \\ 
\ 24: \quad \textbf{end while} \\
\ 25:  \textbf{end for} \\ 
\ 26:  \textbf{Return} ChooseBest(\(T\)) \\ 
\textbf{Output}: Highest-rewarded solution \(p_{1:K}^*\) \\ 
\hline
\end{tabular}%
}
\caption{Pseudocode for LE-MCTS.}
\label{tab:pseudo_code}
\end{table}
\section{Experiments}
\label{sec:experiment}
\begin{table*}[t]
\centering
\resizebox{\textwidth}{!}{%
\begin{tabular}{lllcccccc}
\hline
\textbf{Category} & \textbf{Base LLM} & \textbf{Method} & \textbf{GSM8K} & \textbf{MATH} & \textbf{SVAMP} & \textbf{ASDiv} & \textbf{MQA} & \textbf{Average} \\ \hline
\multirow{16}{*}{\textbf{Single LLM}} & \multirow{4}{*}{LLaMA-3} & Greedy & 69.4 & 12.0 & 81.2 & 77.9 & 21.4 & 52.4 \\ 
 &  & SC & 69.3 & 11.8 & 79.5 & 76.4 & 18.9 & 51.2 \\ 
 &  & BS & 74.2 & 19.0 & 81.0 & 79.8 & 21.7 & 55.1 \\ 
 &  & BoN & 74.6 & 13.4 & 83.3 & 77.7 & 16.6 & 53.1 \\ 
 \cline{2-9}
 & \multirow{4}{*}{Gemma-2} & Greedy & 80.9 & 40.4 & 69.2 & 65.6 & 27.9 & 56.8 \\ 
 &  & SC & 80.6 & 39.4 & 68.1 & 66.2 & 27.0 & 56.3 \\ 
 &  & BS & 81.4 & 40.8 & 67.3 & 67.2 & 28.6 & 57.1 \\ 
 &  & BoN & \underline{82.7} & \underline{41.6} & 73.2 & 69.5 & 29.1 & 59.2 \\ \cline{2-9}
 & \multirow{4}{*}{DeepSeek-Math} & Greedy & 46.6 & 28.6 & 64.0 & 70.6 & 63.8 & 54.7 \\ 
 &  & SC & 47.1 & 27.8 & 60.2 & 68.0 & 60.9 & 52.8 \\ 
 &  & BS & 52.4 & 29.0 & 60.1 & 67.5 & \underline{66.8} & 55.2 \\ 
 &  & BoN & 65.9 & 35.0 & 73.0 & 83.5 & 66.1 & 64.7 \\ 
 \cline{2-9}
 & \multirow{4}{*}{Rho-Math} & Greedy & 67.6 & 29.6 & 76.6 & 77.8 & 55.8 & 61.5 \\ 
 &  & SC & 66.9 & 28.2 & 74.2 & 77.3 & 57.5 & 60.8 \\ 
 &  & BS & 69.9 & 28.8 & 77.7 & 81.1 & 58.2 & 63.1 \\ 
 &  & BoN & 74.8 & 34.6 & 79.8 & 82.2 & 61.6 & 66.6 \\ \hline
\multirow{6}{*}{\textbf{Ensemble}} & Top-3 & BoE & 80.0 & 36.0 & \textbf{84.5} & \underline{83.8} & 65.1 & \underline{69.9} \\ 
 & Top-3 & EBS & 66.7 & 41.0 & 80.8 & 78.2 & 64.0 & 66.1 \\ 
 & All & Blender \dag & 51.9 & 1.4 & 71.3 & 69.0 & 21.9 & 43.1 \\ 
 & Top-3 & MoA \dag & 42.5 & 22.2 & 44.3 & 47.4 & 60.4 & 43.4\\
 & All & EVA \dag & 66.3 & 26.0 & 73.8 & 81.4 & 54.6 & 60.4 \\ 
 & Top-3 & Ours & \textbf{84.1 (+1.4)} & \textbf{45.2 (+3.6)} & \underline{84.0} (-0.5) & \textbf{84.4 (+0.6)} & \textbf{71.1 (+4.3)} & \textbf{73.8 (+3.9)} \\ \hline
\end{tabular}%
}
\caption{\textbf{Summary of main results.} We measure the accuracy on the test set of five math reasoning benchmarks. We also report the average of the performances on five datasets in the rightmost column. We highlight the best model in \textbf{bold} and the second-best model with an \underline{underline}, respectively. \dag: we reuse the official code for experiments.}
\label{tab:main_result}
\end{table*} 

\subsection{Experimental Settings}
\label{text:experiment:setting}
\par \noindent \textbf{Datasets and Evaluation} \; We conducted our experiments using five widely used math reasoning datasets: GSM8K~\citep{cobbe-etal-2021-gsm8k}, MATH~\citep{hendrycks-etal-2021-math}, SVAMP~\cite{patel-etal-2021-nlp}, ASDiv~\cite{miao-etal-2020-diverse}, and MQA~\cite{amini-etal-2019-mqa}.
For the MATH dataset, we used the MATH500 subset to avoid data leakage, which is identical to the test set used by~\citet{lightman-etal-2023-prm800k}.
For the other datasets, we used the entire test set for evaluation.
We used the math-evaluation-harness library~\cite{gou-etal-2024-harness} to ensure consistency and comparability with existing work, following the evaluation framework of DeepSeek-Math~\citep{shao-etal-2024-deepseek}.
We used the in-context examples provided by the math-evaluation-harness for few-shot chain-of-thought (CoT) prompting and report accuracy as the performance metric.

\medskip
\par \noindent \textbf{Baselines} \; 
We benchmarked our method against both single LM decoding algorithms and LM ensemble approaches. 
We evaluated two reward-free algorithms, greedy decoding and self-consistency (SC)~\citep{wang-etal-2023-selfconsistency}, as well as two process reward-guided algorithms, Best-of-N (BoN)~\citep{lightman-etal-2023-prm800k} and Beam Search (BS)~\citep{yu-etal-2024-ovm}, as baselines for single LMs.
We also compared our method with the token-level ensemble approach EVA~\citep{xu-etal-2024-bridging}, as well as the output-level ensemble methods LLM-Blender~\citep{jiang-etal-2023-llm} and MoA~\citep{wang-etal-2024-mixture}.

Furthermore, we propose two novel variations of process reward-guided decoding algorithms by leveraging multiple LLMs. 
Specifically, Best-of-Ensemble (BoE) selects the highest-rewarded one among the complete solutions generated by various LLMs as the final output.
Ensemble Beam Search (EBS) generates candidate reasoning steps using multiple LLMs instead of a single LLM and performs beam search.
Further details of the baselines are provided in Appendix~\ref{text:appx:baselines}.
\medskip
\par \noindent \textbf{Implementation Details} \; We considered two general domain LLMs, LLaMA-3 8B~\citep{dubey-etal-2024-llama3} and Gemma-2 9B~\citep{gemma-etal-2024-gemma2}, and two math LLMs, DeepSeek-Math 7B~\citep{shao-etal-2024-deepseek} and Rho-Math 7B~\citep{lin-etal-2024-rho}, as base models for the model ensemble. 
We set \( n_{\text{iter}}=200 \) and perform LM random selection without replacement for all experiments, except when the specific hyperparameter setting is mentioned.
After running LE-MCTS on each example, we extracted all trajectories in the MCTS tree that reached the terminal node. 
We then ranked these trajectories using PRM and selected the top-ranked trajectory as the final output. 
We tested two sets of base models for ensemble methods, Top-3 and All, and report the one with the better average performance as the main result. 
Specifically, for Top-3, a distinct set of LLMs was used for each dataset, while for All, we ensembled all four base LLMs. 
Full experimental results and further details of the base model selection for Top-3 are provided in Appendix~\ref{text:appx:implementation}.

\subsection{Main Results}
\label{text:experiment:main_results}
We report the main results on math benchmarks in Table~\ref{tab:main_result}.
\medskip
\par \noindent \textbf{LE-MCTS outperforms or matches existing approaches.} \;
The results show that LE-MCTS matches the performance of the best model on SVAMP and outperforms all other models in the remaining datasets. 
Specifically, the average performance across the five tasks is 3.9\% higher than the second-best model, BoE. 
This suggests that LE-MCTS is a versatile language model ensemble framework for complex reasoning, particularly in mathematical problem-solving.
\medskip
\par \noindent \textbf{LE-MCTS is especially good at complex reasoning.} \;
LE-MCTS significantly outperforms the second-best model on MATH and MQA. 
MATH consists of competition-level math problems, while MQA includes GRE and GMAT exam questions, both of which require more diverse and complex reasoning skills than other grade-school math datasets. 
These challenging problems often involve longer reasoning steps because of the need for in-depth reasoning.
\begin{figure}[t]
    \centering
    \includegraphics[width=\columnwidth]{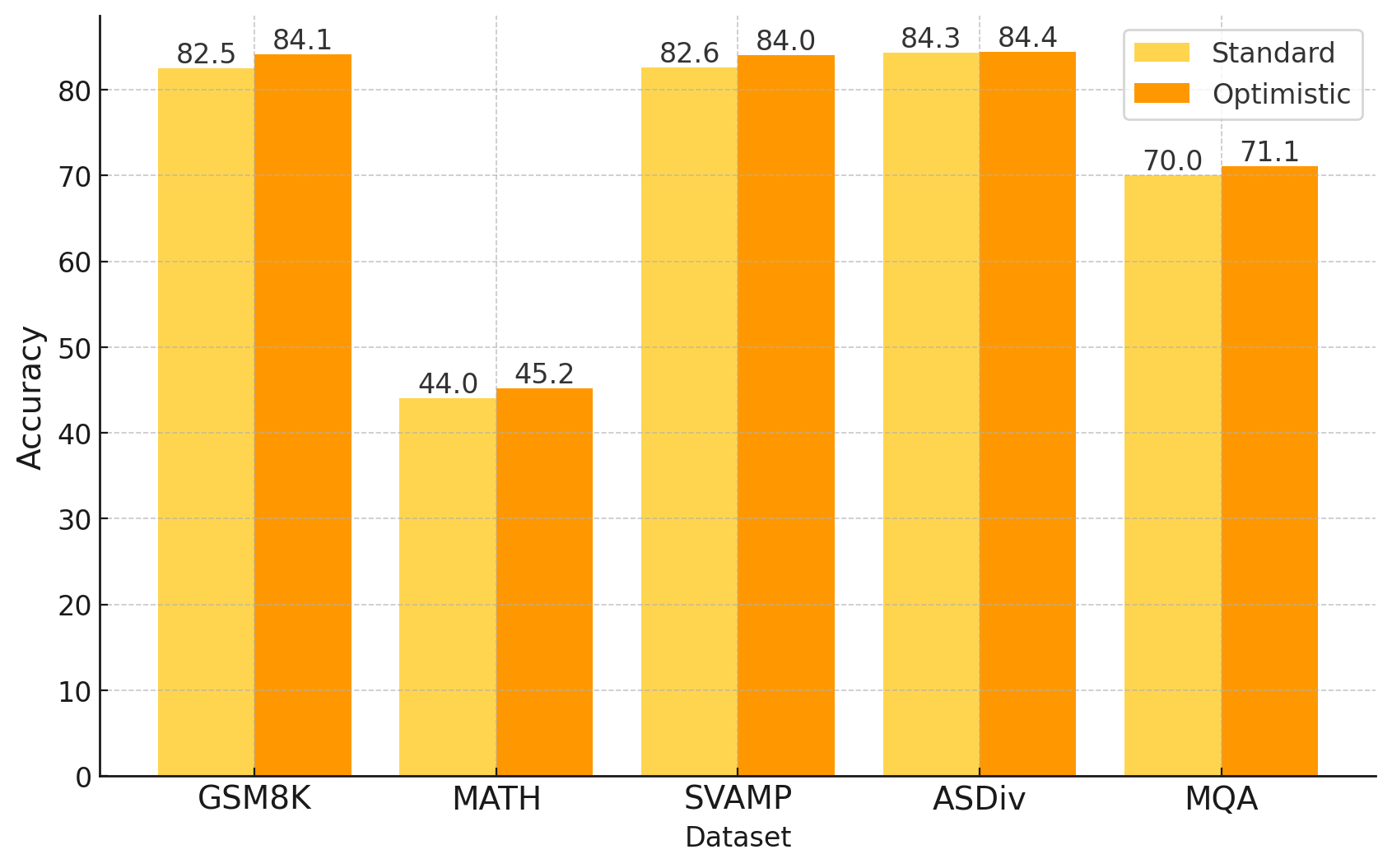}
    \caption{Ablation study on value backpropagation strategies.}
    \label{fig:exp:ablation}
\end{figure}

What distinguishes LE-MCTS from other approaches is its ability to prioritize in-depth reasoning. 
For example, the full-expansion criterion guides LE-MCTS to explore deeper reasoning steps rather than expanding the breadth of early-stage reasoning. 
Moreover, the small UCT constant \( C \) encourages the selection of high-value nodes over underexplored nodes. 
When combined with optimistic backpropagation, this setting also makes LE-MCTS favor deeper exploration.
In contrast, BoE and EBS rely on nucleus sampling to generate reasoning steps, which does not guarantee deeper or more thorough reasoning.
\medskip
\par \noindent \textbf{Process-level ensembles outperform token- and output-level approaches in math reasoning.} \;
Token-level and output-level ensemble approaches perform significantly worse than process-level ensemble methods. 
In some cases, they even perform worse than the weakest single LLM baseline.
For instance, LLM-Blender performs poorly on MATH, achieving an accuracy of only 1.4\%. 
We believe that GenFuser, which aggregates outputs from multiple LLMs to generate the final result, has limited capacity for comprehending complex reasoning.
In contrast, MoA employs a larger language model for output fusion and utilizes prompt engineering to enhance aggregation. 
Consequently, its accuracy on MATH surpasses that of LLM-Blender, though it still falls short of process-level ensemble methods. 

EVA demonstrates better average performance than output-level ensemble baselines.
However, its vocabulary alignment requires matching embedding dimensions across the base models, which limits its applicability. 
As a result, we cannot use Gemma-2 as the base model for EVA, even though it achieves the highest performance among the base LLMs on GSM8K and MATH. 
In contrast, process-based algorithms, including ours, are not constrained by architecture or vocabulary, ensuring broad applicability across various LLMs.

\subsection{Analysis}
\label{text:experiment:cost_analysis}
\medskip
\par \noindent \textbf{Backpropagation Strategy} \;
As an ablation study, we compared LE-MCTS with standard value backpropagation to LE-MCTS with optimistic value backpropagation. 
The results in Fig.~\ref{fig:exp:ablation} show that the optimistic backpropagation consistently improves performance across all datasets, with increases ranging from 0.1\% to 1.6\%.
Moreover, as shown in Appendix~\ref{text:appx:r_dist}, we observed higher average rewards for leaf nodes when using optimistic backpropagation.
This occurs because optimistic backpropagation updates only the values of the highest-value nodes through the tree, thereby neglecting low-value nodes during the search. 
As PRM more accurately estimates the process rewards, the performance gap between optimistic and standard strategies is expected to widen.
\medskip
\begin{table}[t]
\centering
\resizebox{\columnwidth}{!}{%
\begin{tabular}{cccccc}
\hline
\textbf{\( C \)} & \textbf{GSM8K} & \textbf{MATH} & \textbf{SVAMP} & \textbf{ASDiv} & \textbf{MQA} \\
\hline
 0.5    & 81.7 & \textbf{45.2} & 82.7 & 84.2 & \textbf{71.1} \\
 1.0    & 83.7 & 43.6 & \textbf{84.0} & \textbf{84.4} & 69.0 \\
 1.414  & \textbf{84.1} & 44.4 & 83.8 & 84.2 & 68.6 \\
\hline
\end{tabular}
}
\caption{Effect of the UCT constant \( C \).}
\label{tab:uct_const}
\end{table}
\par \noindent \textbf{UCT Constant, \( C \)} \;
The constant \( C \) controls the balance between exploration and exploitation in LE-MCTS. 
In our framework, the depth of the node corresponds to the number of reasoning steps.
The experimental results in Appendix~\ref{text:appx:d_dist} show that the average depth of the leaf nodes tends to increase as \( C \) decreases.
We hypothesize that simpler math problems, such as grade-school math word problems (\textit{e.g.}, SVAMP and ASDiv), do not require long reasoning chains and benefit from a higher \( C \), as it facilitates more effective reasoning in the early steps, such as identifying mathematical variables. 
Conversely, more challenging math problems, such as those in MATH and MQA, require competition-level reasoning skills. 
In these cases, a lower \( C \) is advantageous, as it encourages a deeper exploration of reasoning paths instead of focusing on the early steps.

To test our hypothesis, we evaluated LE-MCTS with \( C \in \{0.5, 1.0, 1.414\} \)\footnote{We follow the standard practice~\citep{kocsis-etal-2006-uct, browne-etal-2012-survey} to choose the value \(\sqrt{2}\), along with smaller values of 0.5 and 1.0 for greater exploitation.}, and the results are reported in Table~\ref{tab:uct_const}. The results support our hypothesis that \( C = 0.5 \) yields the best performance on MATH and MQA, while \( C = 1.0 \) and \( C = 1.414 \) outperform \( C = 0.5 \) on GSM8K, SVAMP, and ASDiv. 
Our experiment offers a guide for choosing an appropriate \( C \) value for previously unseen problems. 
For problems with a complexity level comparable to simple grade-school math word problems, a \( C \) value greater than 1 is recommended to facilitate effective exploration of early-stage reasoning steps. 
In contrast, for more complex problems, a \( C \) value below 1 is preferable, as it encourages a more thorough investigation of the in-depth reasoning paths.
\medskip
\begin{table}[t]
\centering
\resizebox{\columnwidth}{!}{%
\begin{tabular}{cccccc}
\hline
\textbf{\(n_{iter}\)} & \textbf{GSM8K} & \textbf{MATH} & \textbf{SVAMP} & \textbf{ASDiv} & \textbf{MQA} \\
\hline
10  & 79.8 & 43.8 & 82.9 & 82.4 & 65.7 \\
25  & 81.0 & 43.4 & 82.9 & 83.5 & 65.5 \\
50  & 81.5 & 44.4 & 82.7 & 83.2 & 68.9 \\
100 & 82.9 & 45.2 & 83.1 & 83.5 & 68.2 \\
200 & \textbf{84.1} & \textbf{45.2} & \textbf{84.0} & \textbf{84.4} & \textbf{71.1} \\
\hline
\end{tabular}
}
\caption{Effect of the maximum number of MCTS iterations \( n_{iter} \).}
\label{tab:n_iter}
\end{table}
\par \noindent \textbf{Maximum Number of MCTS Iterations, \(n_{iter}\)} \;
The hyperparameter \(n_{iter}\) plays a crucial role in determining the balance between performance and computational cost.
A large \(n_{iter}\) increases the execution time of the algorithm, while a too small \(n_{iter}\) may be insufficient to find an optimal reasoning path.
To investigate the effect of \(n_{iter}\) on performance, we varied \(n_{iter}\) over the set \(\{10, 25, 50, 100, 200\}\) and report the results in Table~\ref{tab:n_iter}.
The results show that performance improves as \(n_{iter}\) increases from 10 to 200, suggesting that we can find a better reasoning path as we run more MCTS iterations.
However, this improvement varies across datasets, and performance generally saturates beyond 200 iterations.
These results highlight the importance of selecting an appropriate \(n_{iter}\) to balance performance with computational cost. 
\medskip
\begin{table}[t]
\centering
\resizebox{\columnwidth}{!}{%
\begin{tabular}{l|cc|cc}
\hline
\multirow{2}{*}{\textbf{Method}} & \multicolumn{2}{c|}{\textbf{ASDiv}} & \multicolumn{2}{c}{\textbf{MATH}} \\
\cline{2-3}
\cline{4-5}
 & VRAM ($\downarrow$) & min/ex ($\downarrow$) & VRAM ($\downarrow$) & min/ex ($\downarrow$) \\
\hline
  BoE       & 76.7 & 17.6 & 64.8 & 71.1 \\
  EBS       & 79.2 & 12.3 & 71.8 & 47.2 \\
  Blender   & 70.4 & 22.2 & 66.5 & 59.1 \\
  MoA       & 67.4 & 84.4 & 79.8 & 93.7 \\
  EVA       & 69.9 & 92.2 & 70.3 & 480.2 \\
  $\text{Ours}_{n_{iter}=25}$ & 76.7 & 34.6 & 64.6 & 129.1 \\
  $\text{Ours}_{n_{iter}=200}$ & 77.4 & 112.2 & 71.0 & 342.2 \\
\hline
\end{tabular}
}
\caption{\textbf{Efficiency analysis.} We compare the efficiency of our method with existing ensemble approaches based on peak VRAM usage and throughput. Throughput is measured as the average time per example, reported in minutes per example (min/ex). VRAM usage is quantified as the maximum value observed during inference, expressed in gigabytes (GB). For both metrics, lower values indicate higher efficiency.}
\label{tab:efficiency}
\end{table}
\par \noindent \textbf{Efficiency Analysis} \;
We follow \citet{dehghani-etal-2022-efficiency} to compare the efficiency of the proposed method with existing ensemble approaches. 
Specifically, we analyze efficiency along two axes: GPU memory consumption and throughput. 
Throughput is defined as the average time required to process each example. 
We use peak VRAM usage during inference as the memory consumption metric.
We choose VRAM usage as an efficiency metric because the number of floating-point operations (FLOPs), although commonly used, is difficult to compare fairly across different decoding algorithms. 
In contrast, VRAM usage serves as an indirect indicator of the number of parameters and the input sequence length involved in the decoding process, which are the most important factors for measuring computational cost in Transformer architectures.

For a fair comparison, we measured VRAM usage and throughput on a single NVIDIA H100 80GB GPU, using the same CUDA and PyTorch versions. 
We evaluated these efficiency metrics on the most complex dataset, MATH, and the simplest dataset, ASDiv.
The results in Table~\ref{tab:efficiency} show that EBS is the most efficient algorithm in terms of throughput.
For peak VRAM usage, MoA and \(\text{LE-MCTS}_{n_{iter}=25}\) are the most efficient algorithms on ASDiv and MATH, respectively.
Both output- and token-level ensemble methods show high throughput in MATH, while LLM-Blender and MoA suffer from the long input sequence length of the aggregator, and EVA lacks optimization for generating long sequence in its implementation.

Our LE-MCTS with \(n_{iter}=200\) is more resource-intensive than most existing algorithms but achieves better performance.
While LE-MCTS offers significant performance improvements on complex problems, its advantages on simpler tasks are marginal. 
In ASDiv, LE-MCTS is less efficient than BoE and EBS in throughput, even when we run a small number of MCTS iterations \(n=25\).
This results underscore that while LE-MCTS's in-depth search is beneficial for solving complex reasoning problems, it may introduce unnecessary computational overhead for straightforward tasks, where EBS and BoE can achieve comparable performance at a lower cost. 
Therefore, LE-MCTS is particularly well-suited for challenging reasoning tasks that require in-depth analysis, where a trade-off between performance and computational cost is acceptable.

\section{Related Work}
\label{sec:relwork}
\subsection{Language Model Ensemble}
\label{text:relwork:llm_ensemble}
Ensemble learning is a widely used approach to improve model performance by integrating multiple weaker models~\citep{lu-etal-2024-lesurvey}. 
In large language models (LLMs), ensemble methods have been typically applied at either the token level or the output level. 
Token-level ensembling~\citep{liu-etal-2021-dexperts,liu-etal-2024-tuning,li-etal-2024-purifying,mavromatis-etal-2024-packllm} has involved merging token logits or probabilities from multiple LLMs, which often requires that the models share the same token vocabulary and model architecture. 
To overcome this limitation, EVA~\citep{xu-etal-2024-bridging} have mapped the output distributions of different LLMs into a unified space through pre-trained mapping. 
In contrast, output-level ensemble methods~\citep{jiang-etal-2023-llm,wang-etal-2024-mixture,izacard-grave-2021-leveraging,ravaut-etal-2022-towards} have combined entire model outputs using an additional fusion model. 
For example, LLM-Blender~\citep{jiang-etal-2023-llm} have introduced a general framework that employs a pair ranker to filter the top K optimal outputs before merging them through a fusion model to generate the final result. 

Our work introduces a novel process-level ensemble method that offers greater flexibility than token-level approaches, as LE-MCTS does not need to match the vocabulary or architecture. 
Moreover, in output-level ensemble methods, the aggregator receives the outputs of multiple LLMs as input, which can cause the input sequence length to exceed the limit it can handle.
In contrast, the maximum input sequence length of LE-MCTS remains the same as when solving a math reasoning problem with a single LLM using greedy decoding. 
Therefore, output-level ensemble methods are not suitable for complex reasoning tasks that require long, in-depth reasoning.

\subsection{Reward-Guided Decoding}
\label{text:relwork:reward_decoding}
Two main types of verifiers for mathematical reasoning problems are the Outcome Reward Model (ORM) and the Process Reward Model (PRM). 
ORM assesses the entire solution, whereas PRM evaluates individual steps, providing more granular feedback. 
Both approaches have shown improvements in mathematical reasoning compared to self-consistency~\citep{wang-etal-2023-selfconsistency}, yet evidence has shown that PRM outperforms ORM~\citep{lightman-etal-2023-prm800k, wang-etal-2024-math}.
Previous studies have used rewards to guide the decoding process at the sentence level~\citep{uesato-etal-2022-oprm, welleck-etal-2022-naturalprover, yao-etal-2024-tree} and the token level~\citep{dathathri-etal-2020-plug, yang-klein-2021-fudge, chaffin-etal-2022-ppl, li-etal-2023-contrastive}. 
ARGS~\citep{khanov-etal-2024-args} has incorporated language model alignment as a reward-guided search. 
\citet{yu-etal-2024-ovm} and \citet{ma-etal-2023-let} have applied process rewards to enhance heuristic search algorithms.
Recently, ReST-MCTS*~\citep{zhang-etal-2024-rest} has also employed process reward-guided MCTS, similar to our approach, to generate high-quality data for self-training.
However, our use of MCTS focuses on ensembling LLMs.
To the best of our knowledge, this is the first attempt to apply process-reward guidance for LLM ensembling.
\section{Conclusion}
\label{sec:conclusion}
In this paper, we introduced a novel framework for process-level ensembling of language models, addressing key limitations of traditional language model ensemble methods. 
We formulated the step-by-step reasoning with an ensemble of language models as a Markov decision process. 
By leveraging an existing process-based reward model and Monte Carlo Tree Search, our approach effectively navigates the unified space of reasoning steps generated by different language models, discovering more accurate solutions.
Extensive experiments on mathematical reasoning benchmarks demonstrate the efficacy of the proposed method, especially in solving complex reasoning problems. 
We believe this approach paves the way for broader process-level ensemble of language models, as it can be applied to any step-by-step reasoning problem where an appropriate process-reward model exists.
\section*{Limitations}
\label{sec:limitation}
Our proposed LE-MCTS framework relies on signals from the process-based reward model to effectively navigate the space of reasoning steps. 
When the PRM fails to compute rewards accurately, LE-MCTS also fails. 
While Math-Shepherd's PRM generally performs well on established mathematical datasets, there is no guarantee that it will be applicable to novel math problems. 
Moreover, PRMs for other complex, step-by-step reasoning tasks remain relatively unexplored.
Therefore, developing robust and generalizable PRMs is crucial for enhancing LE-MCTS, and we believe this is a valuable future direction.

Another limitation of our work is the need to select base models. 
The experimental results in Appendix~\ref{text:appx:implementation} suggest that using weak base models harms the performance of LE-MCTS.
Although we demonstrated that even a small amount of synthetic data can effectively identify weak base models, this approach may not generalize well to other tasks and datasets. 
Nevertheless, selecting base models provides a clear advantage in terms of efficiency. 
By ensembling only a small number of LLMs, LE-MCTS achieves greater efficiency in both VRAM usage and the number of MCTS iterations.
We believe a robust selection algorithm for identifying effective base models is another important future direction.
\bibliography{anthology,custom}
\newpage
\appendix
\section{Base Model Selection}
\label{text:appx:implementation}
\begin{figure*}[t]
    \centering
    \includegraphics[width=\textwidth]{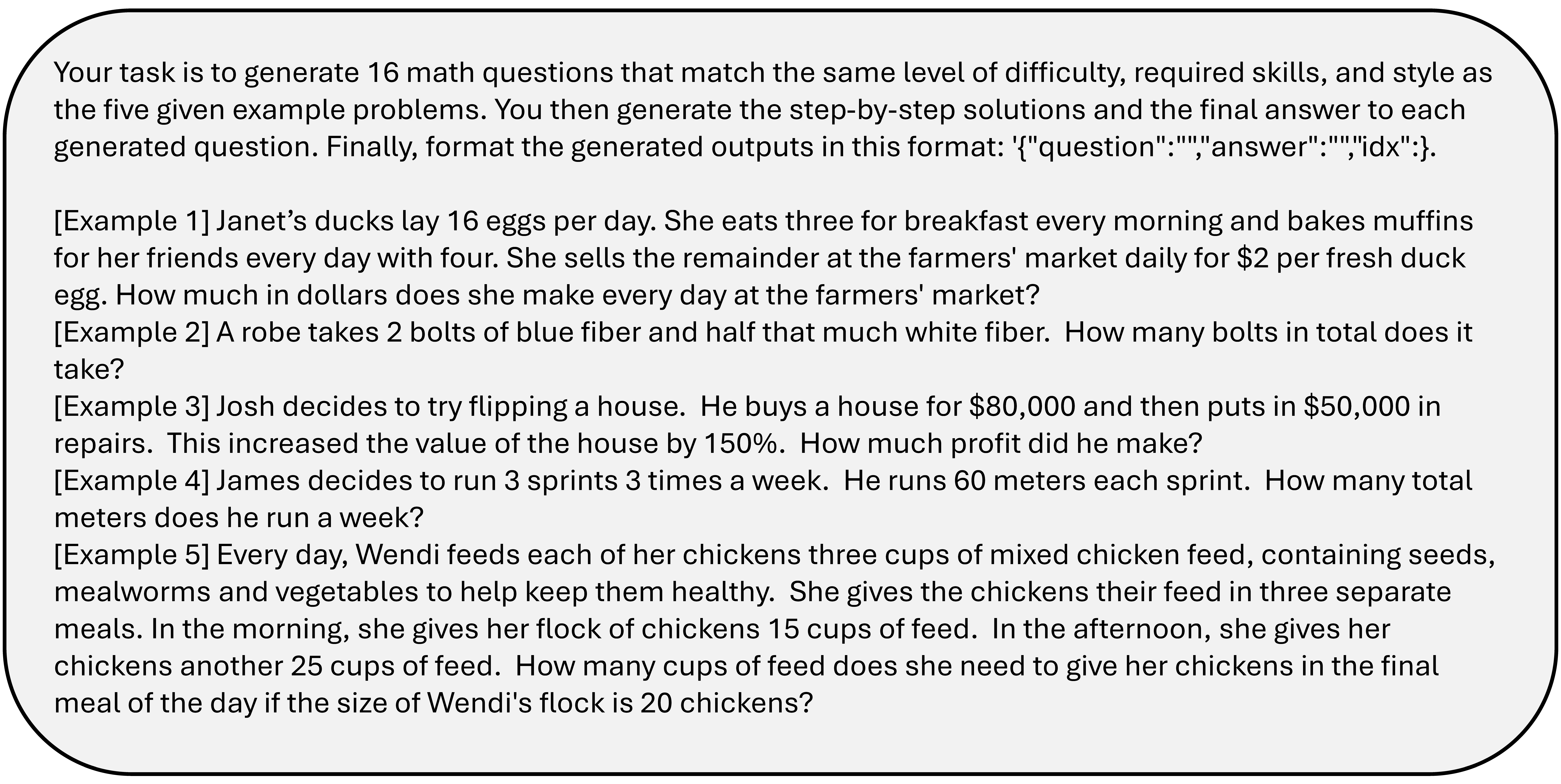}
    \caption{A prompt for generating 16 synthetic examples analogous to those in GSM8K.}
    \label{fig:synprompt_gsm8k}
\end{figure*}

\begin{figure*}[t]
    \centering
    \includegraphics[width=\textwidth]{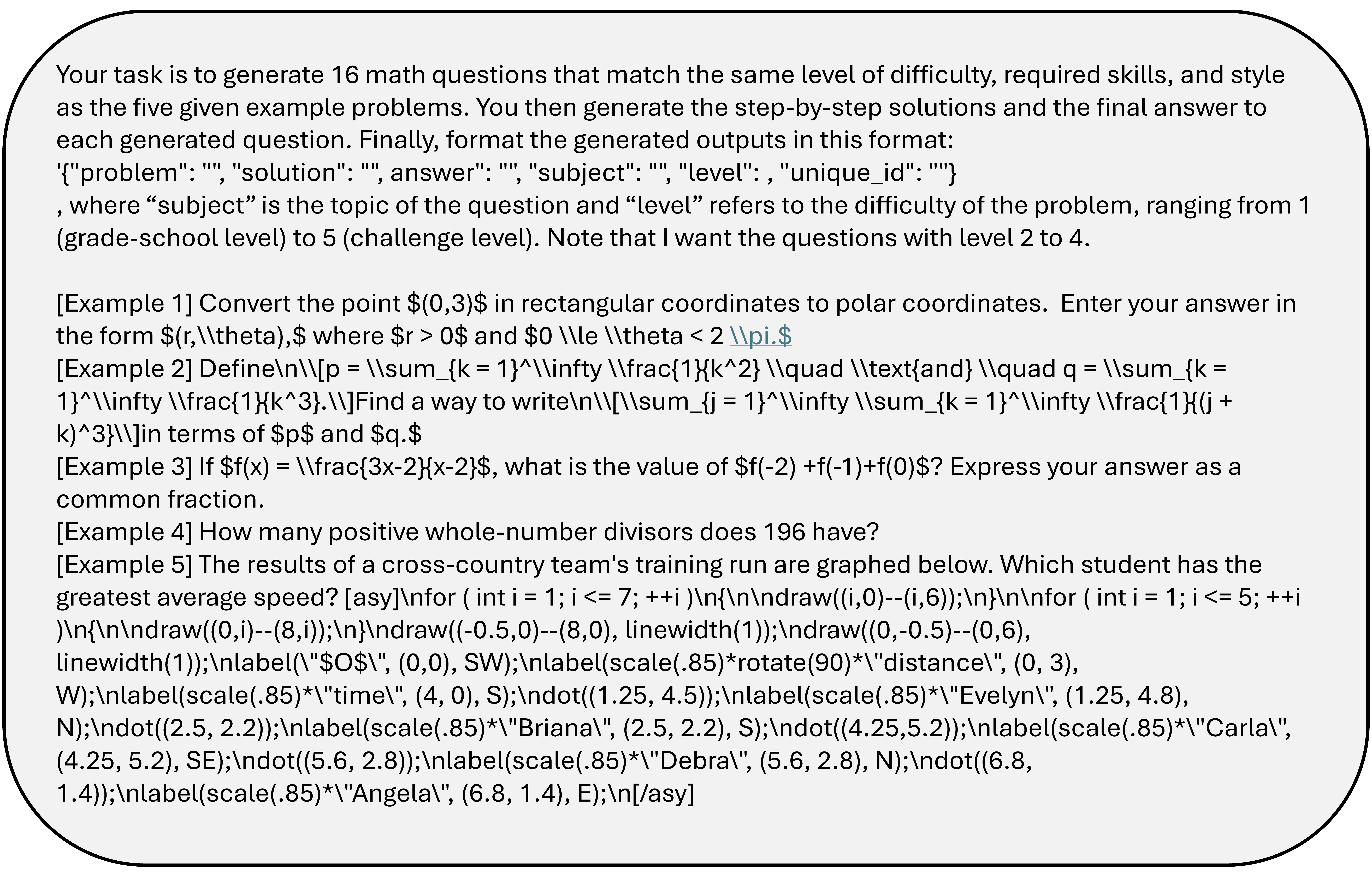}
    \caption{A prompt for generating 16 synthetic examples analogous to those in MATH.}
    \label{fig:synprompt_math}
\end{figure*}

\begin{figure*}[t]
    \centering
    \includegraphics[width=\textwidth]{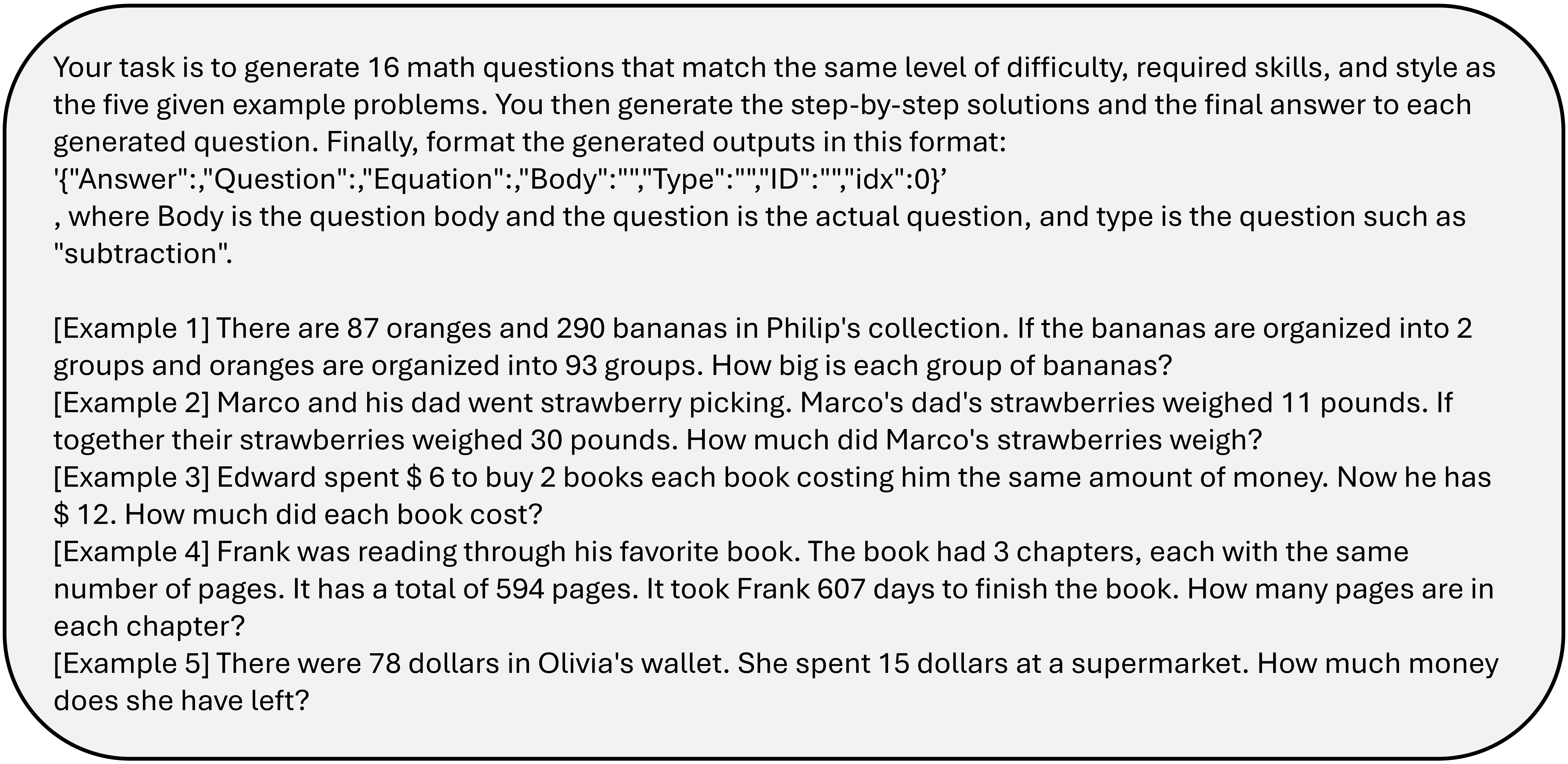}
    \caption{A prompt for generating 16 synthetic examples analogous to those in SVAMP.}
    \label{fig:synprompt_svamp}
\end{figure*}

\begin{figure*}[t]
    \centering
    \includegraphics[width=\textwidth]{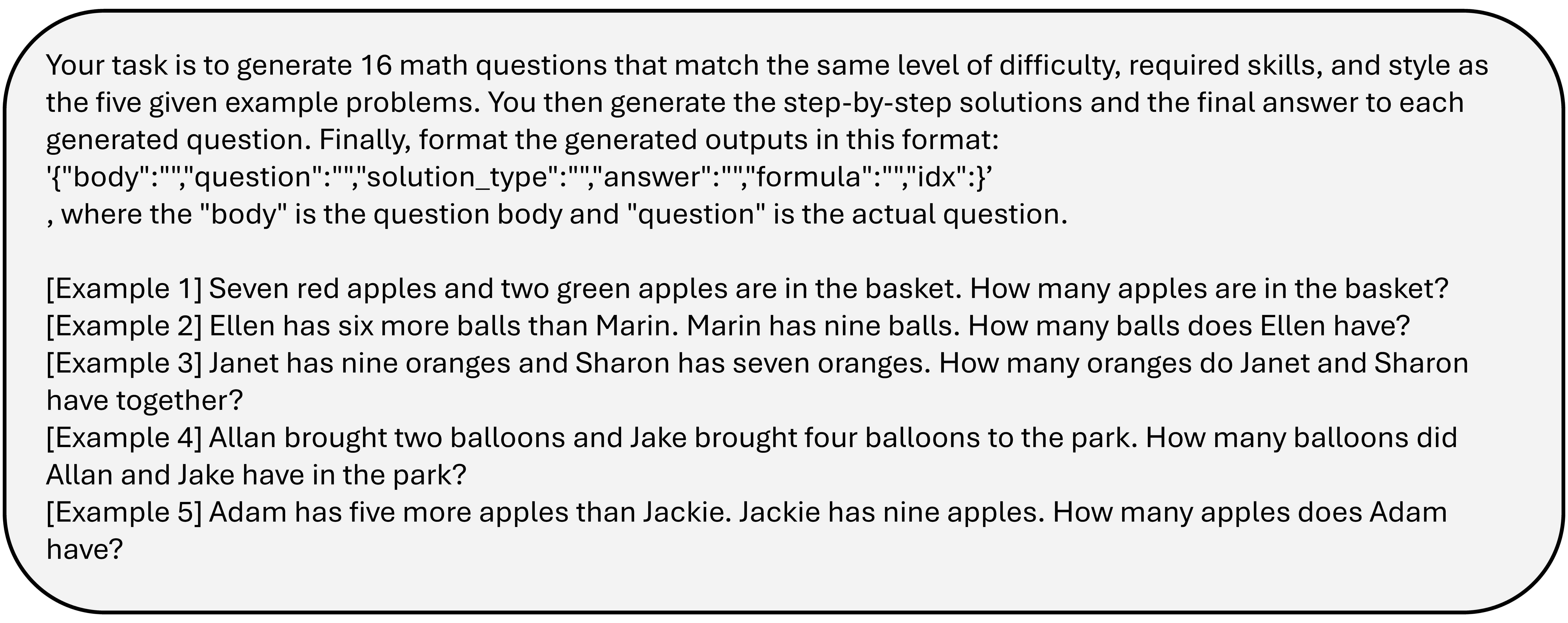}
    \caption{A prompt for generating 16 synthetic examples analogous to those in ASDiv.}
    \label{fig:synprompt_asdiv}
\end{figure*}

\begin{figure*}[t]
    \centering
    \includegraphics[width=\textwidth]{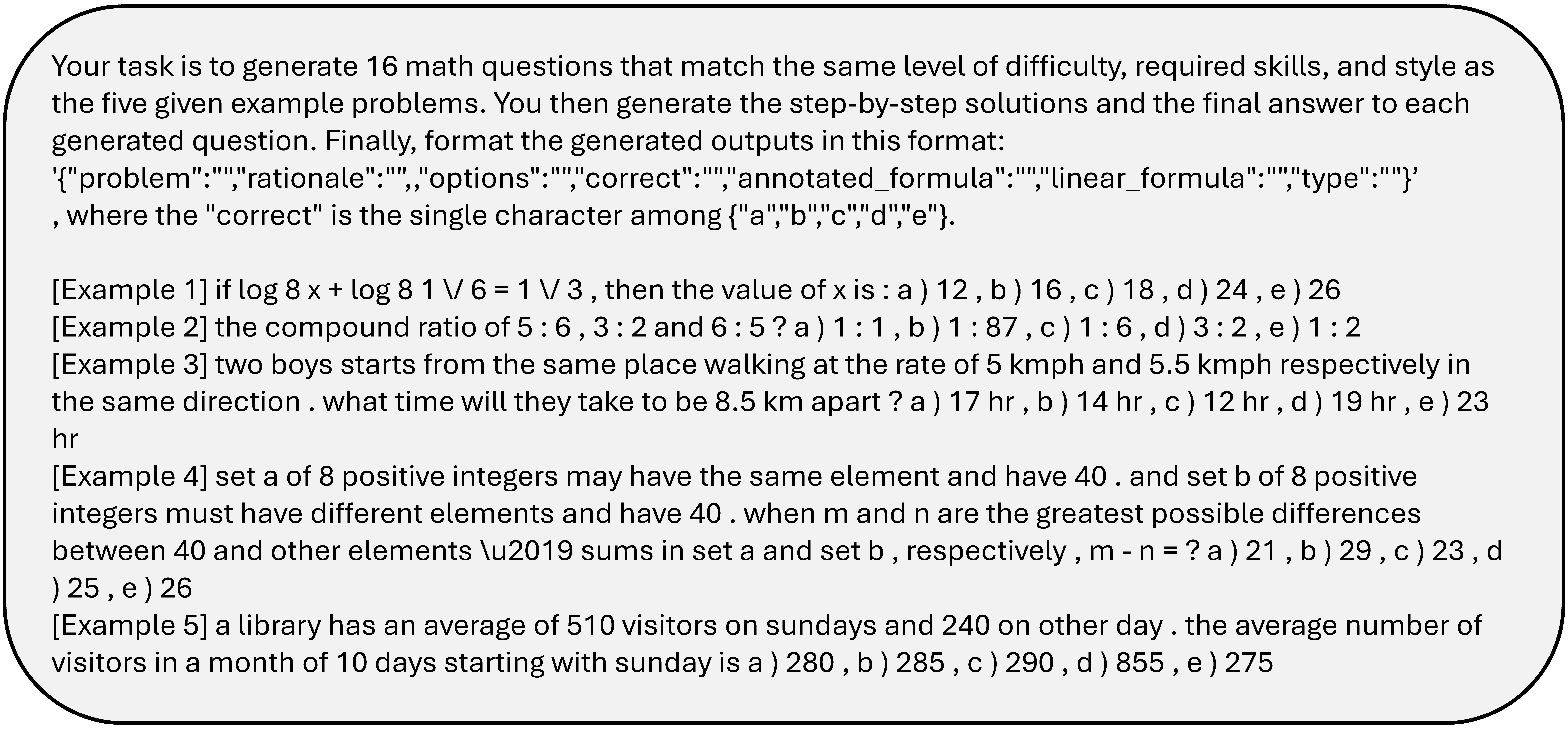}
    \caption{A prompt for generating 16 synthetic examples analogous to those in MQA.}
    \label{fig:synprompt_mathqa}
\end{figure*}

LLMs sometimes exhibit significantly lower performance on certain datasets, such as LLaMA-3 on MATH and MQA. 
We hypothesize that including such LLMs as base models may harm the performance of ensemble frameworks.
However, selecting base models for unseen problems is challenging because we cannot measure their performance in advance.
To address this, we generated synthetic examples that mirror the original math datasets in both difficulty and style, and we used the performance on these synthetic data for base model selection.
Specifically, we instructed GPT-4o to generate 16 synthetic examples for each dataset, as shown in Figs.~\ref{fig:synprompt_gsm8k}-\ref{fig:synprompt_mathqa}. 
We evaluated the greedy decoding performance of each LLM on the synthetic data and selected the top three models as base models for the ensemble. 
The evaluation results on the synthetic data are presented in Table~\ref{tab:appx_synthetic}.

For the selection of top-3 models, we heuristically included high-performing LLMs and excluded low-performing ones. 
For example, although DeepSeek-Math performs worse than the other models on ASDiv, it still demonstrates strong performance. 
Therefore, we included all models as base models.
In contrast, the performance gap between the second and third best models on MQA was substantial, so we selected only Rho-Math and DeepSeek-Math as base models.
For GSM8K, MATH, and SVAMP, we used the top-3 models as base models.
Note that EVA was tested only with All, as it cannot utilize Gemma-2 due to a mismatch in hidden dimensions with the other LLMs.

We present the complete results for the Top-3 and All sets of base models in Table~~\ref{tab:appx_allperformance}. 
Except for LLM-Blender, Top-3 consistently shows better average performance across all ensemble frameworks. 
Moreover, the performance gap between Top-3 and All in LLM-Blender is minimal or zero. 
These results indicate that weak base models indeed harm the performance of ensemble frameworks and that synthetic examples can effectively identify weak base models.

\begin{table}[t]
\centering
\resizebox{\columnwidth}{!}{%
\begin{tabular}{cccccc}
\hline
\textbf{Model} & \textbf{GSM8K} & \textbf{MATH} & \textbf{SVAMP} & \textbf{ASDiv} & \textbf{MQA} \\
\hline
 Rho-Math & 81.2 & 56.2 & 87.5 & 100 & 62.5 \\
 LLaMA-3 & 81.2 & 31.2 & 87.5 & 100 & 25 \\
 Gemma-2 & 87.5 & 81.2 & 37.5 & 100 & 37.5 \\
 DeepSeek-Math & 56.2 & 43.8 & 87.5 & 87.5 & 87.5 \\
\hline
\end{tabular}
}
\caption{Performance on 16 synthetic examples.}
\label{tab:appx_synthetic}
\end{table}
\begin{table}[t]
\centering
\resizebox{\columnwidth}{!}{%
\begin{tabular}{ccccccc}
\hline
\textbf{Method} & \textbf{Base Model} & \textbf{GSM8K} & \textbf{MATH} & \textbf{SVAMP} & \textbf{ASDiv} & \textbf{MQA} \\
\hline
 \multirow{2}{*}{EBS} & Top-3 & 66.7 & 41.0 & 80.8 & 78.2 & 64.0 \\
  & All & 54.6 & 41.8 & 78.8 & 78.2 & 61.3 \\
\hline
 \multirow{2}{*}{BoE} & Top-3 & 80.0 & 36.0 & 84.5 & 83.8 & 65.1 \\
  & All & 79.4 & 29.0 & 85.1 & 83.8 & 51.2 \\
\hline
 \multirow{2}{*}{MoA} & Top-3 & 42.5 & 22.2 & 44.3 & 47.4 & 60.4 \\
  & All & 43.3 & 10.4 & 48.4 & 47.4 & 61.2 \\
\hline
 \multirow{2}{*}{Blender} & Top-3 & 49.9 & 1.4 & 69.0 & 69.0 & 21.9 \\
  & All & 51.9 & 1.4 & 71.3 & 69.0 & 21.9 \\
\hline
 \multirow{2}{*}{LE-MCTS} & Top-3 & 84.1 & 45.2 & 84.0 & 84.4 & 71.1 \\
  & All & 84.2 & 40.0 & 79.7 & 84.4 & 67.5 \\
\hline
\end{tabular}
}
\caption{Full experimental results of language model ensemble approaches with two base model configurations: Top-3 and All.}
\label{tab:appx_allperformance}
\end{table}

\section{Average Reward Distribution of Leaf Nodes}
\label{text:appx:r_dist}
\begin{figure*}[t]
    \centering
    \includegraphics[width=\textwidth]{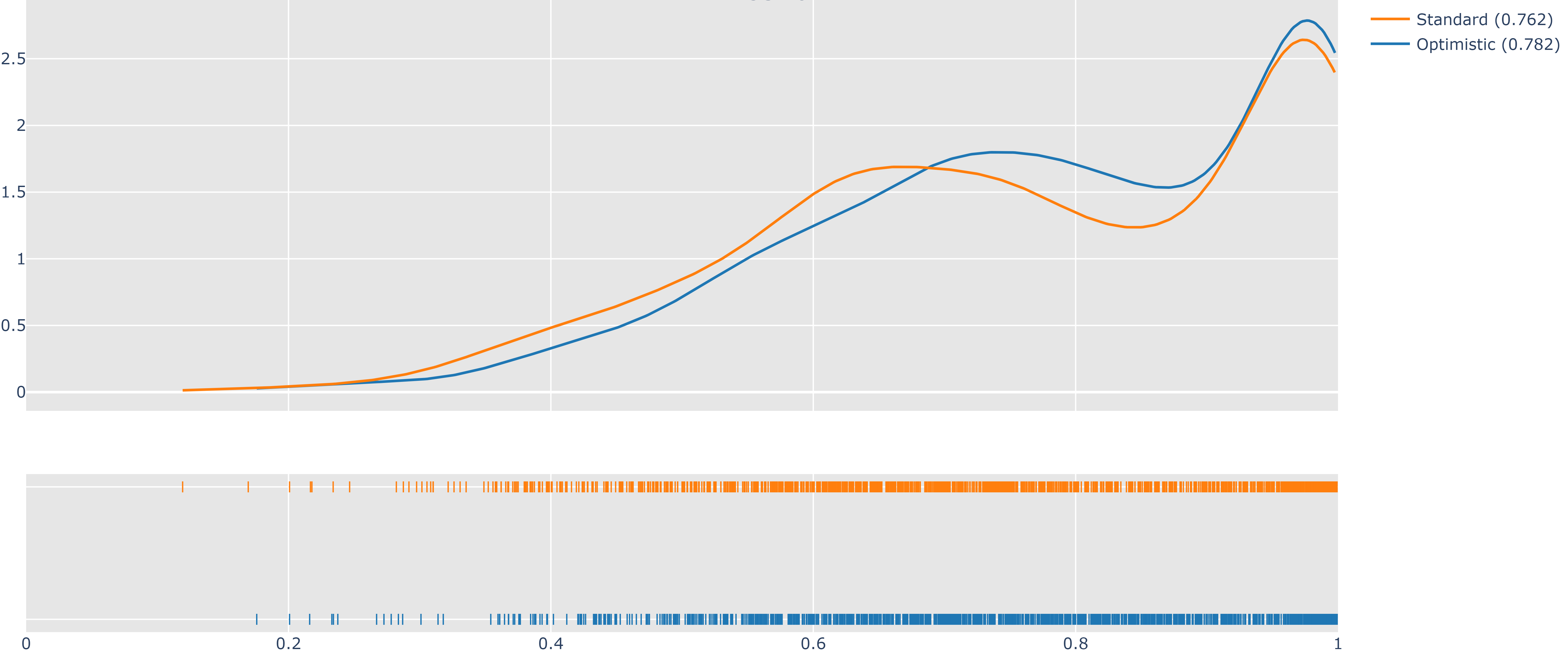}
    \caption{The distribution of the average reward for leaf nodes per example in GSM8K.}
    \label{fig:r_dist_gsm8k}
\end{figure*}

\begin{figure*}[t]
    \centering
    \includegraphics[width=\textwidth]{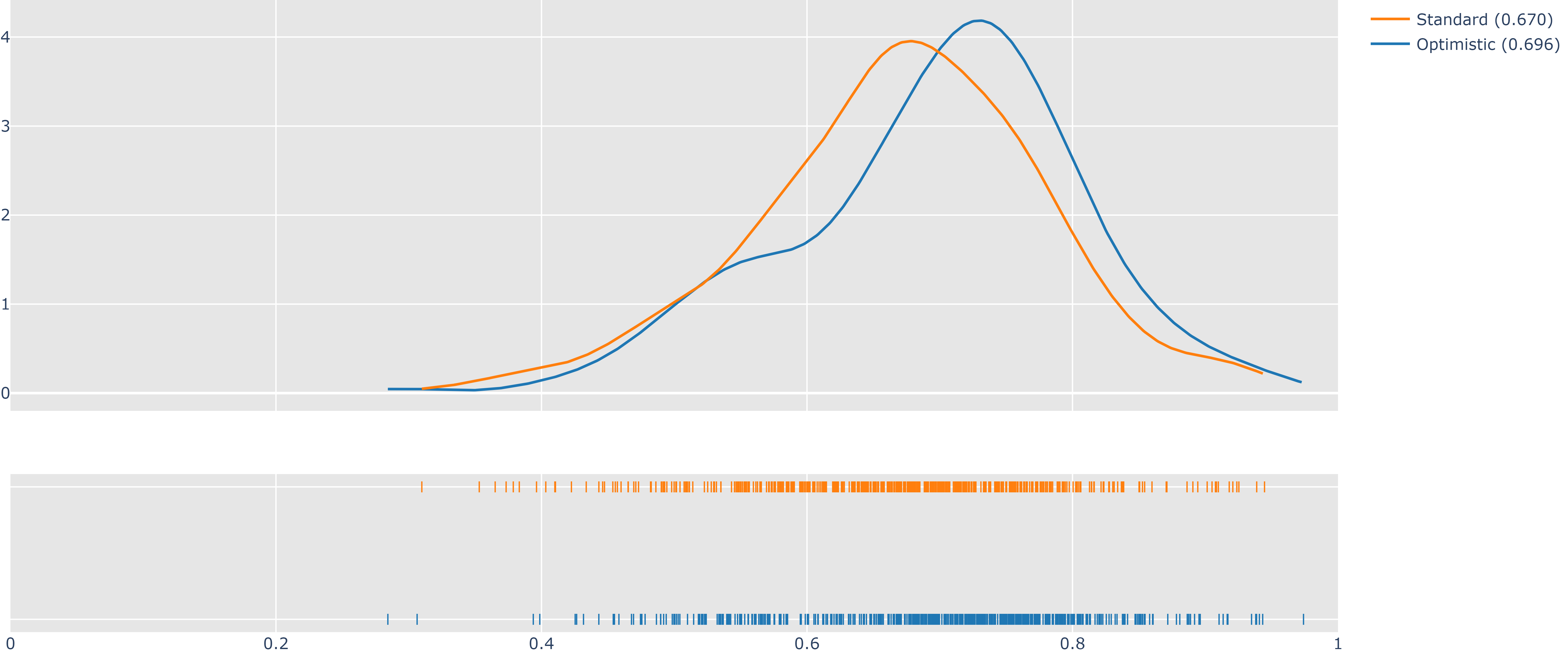}
    \caption{The distribution of the average reward for leaf nodes per example in MATH.}
    \label{fig:r_dist_math}
\end{figure*}

\begin{figure*}[t]
    \centering
    \includegraphics[width=\textwidth]{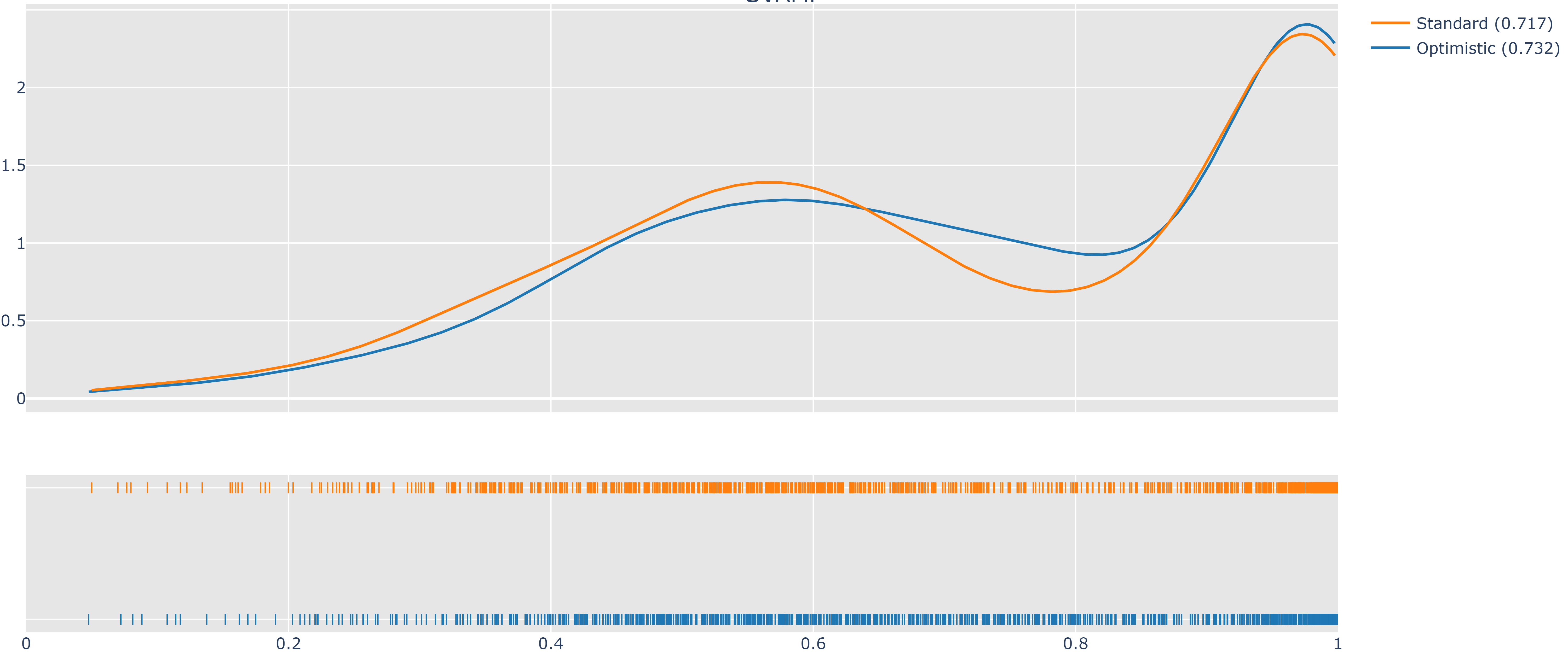}
    \caption{The distribution of the average reward for leaf nodes per example in SVAMP.}
    \label{fig:r_dist_svamp}
\end{figure*}

\begin{figure*}[t]
    \centering
    \includegraphics[width=\textwidth]{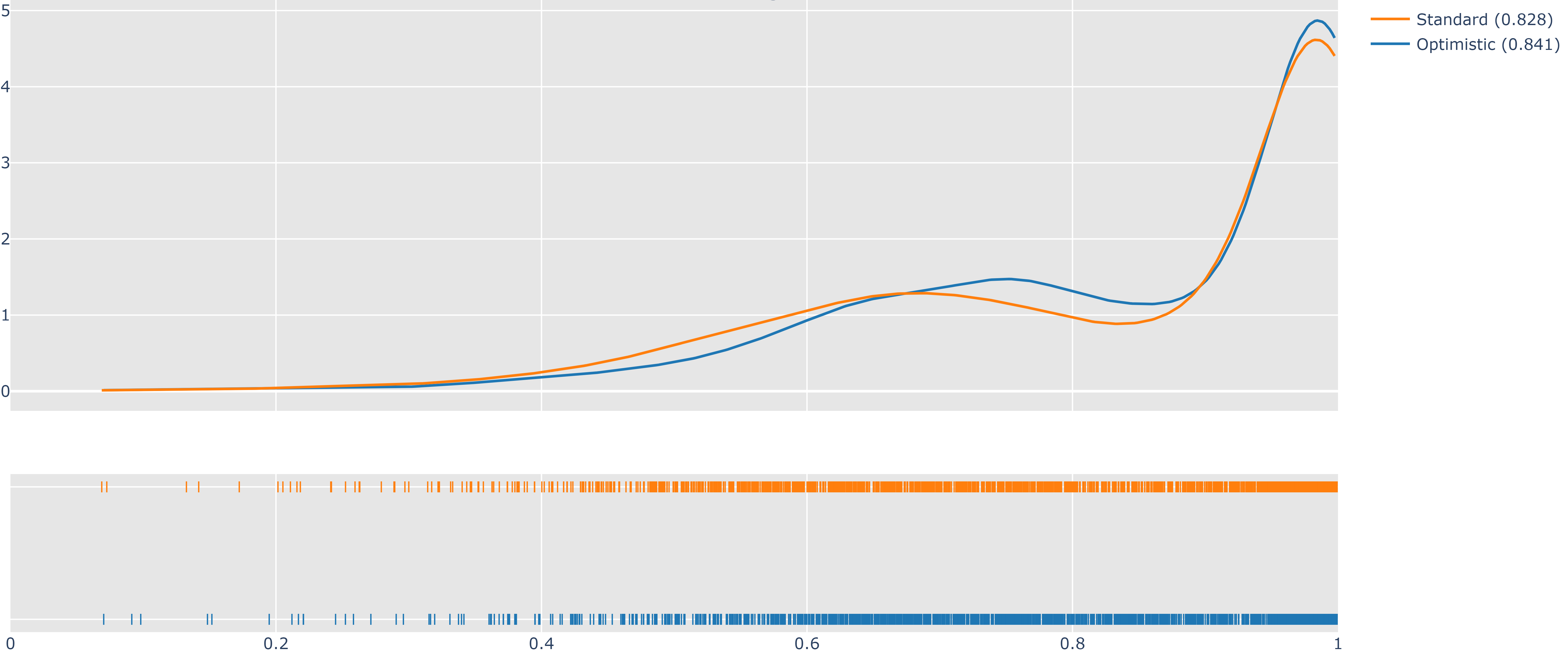}
    \caption{The distribution of the average reward for leaf nodes per example in ASDiv.}
    \label{fig:r_dist_asdiv}
\end{figure*}

\begin{figure*}[t]
    \centering
    \includegraphics[width=\textwidth]{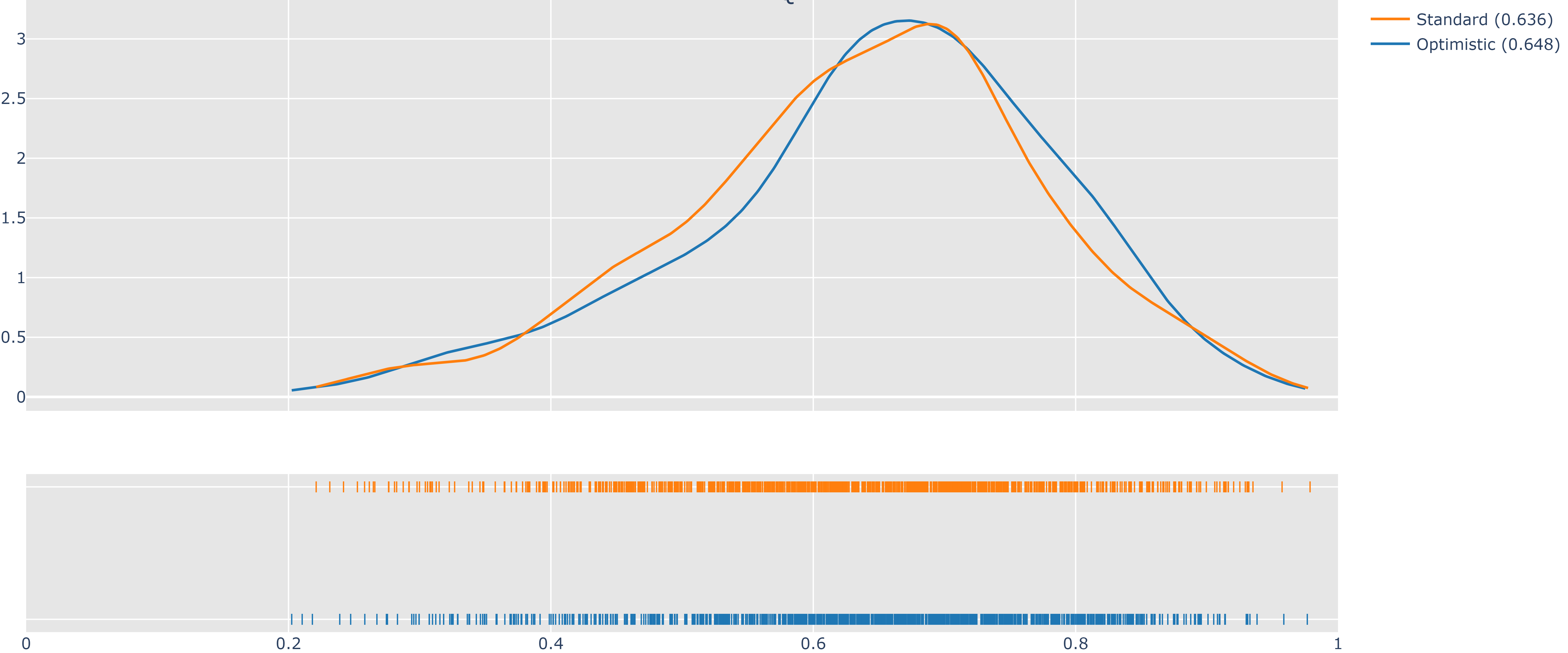}
    \caption{The distribution of the average reward for leaf nodes per example in MQA.}
    \label{fig:r_dist_mathqa}
\end{figure*}

We measured the average reward received by leaf nodes in MCTS for optimistic and standard backpropagation strategies. 
Specifically, we first measured the process rewards of all leaf nodes in the tree, then computed the average per sample and reported the results in the KDE plots Figs.~\ref{fig:r_dist_gsm8k}-\ref{fig:r_dist_mathqa}.
The value in parentheses represents the mean of all points in the plot. 

For complex math problems such as MATH and MQA, leaf nodes received relatively lower rewards compared to simpler problems. 
For easy math problems, like ASDiv, the distribution is left-skewed, with the probability mass concentrated in the range [0.8, 1.0].
Leaf nodes consistently received higher rewards when using optimistic value backpropagation, regardless of the dataset.
These results indicate that optimistic backpropagation is more effective than standard backpropagation in discovering high-reward reasoning trajectories. 
Furthermore, LE-MCTS's preference for high-reward reasoning paths guarantees performance improvements, particularly when paired with a better PRM.

\section{Average Depth Distribution of Leaf Nodes}
\label{text:appx:d_dist}
\begin{figure*}[t]
    \centering
    \includegraphics[width=\textwidth]{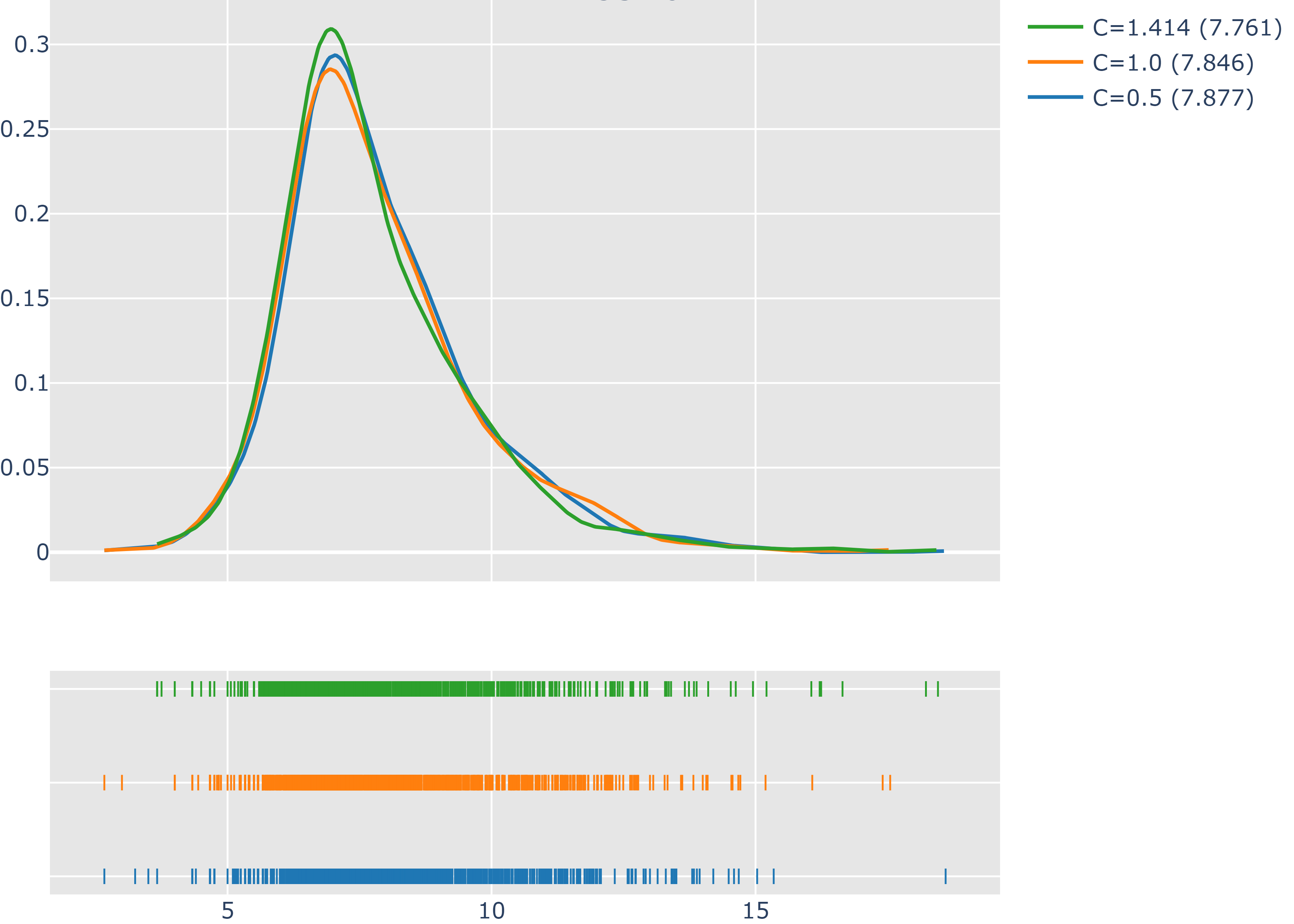}
    \caption{The distribution of the average depth of leaf nodes per example in GSM8K.}
    \label{fig:d_dist_gsm8k}
\end{figure*}

\begin{figure*}[t]
    \centering
    \includegraphics[width=\textwidth]{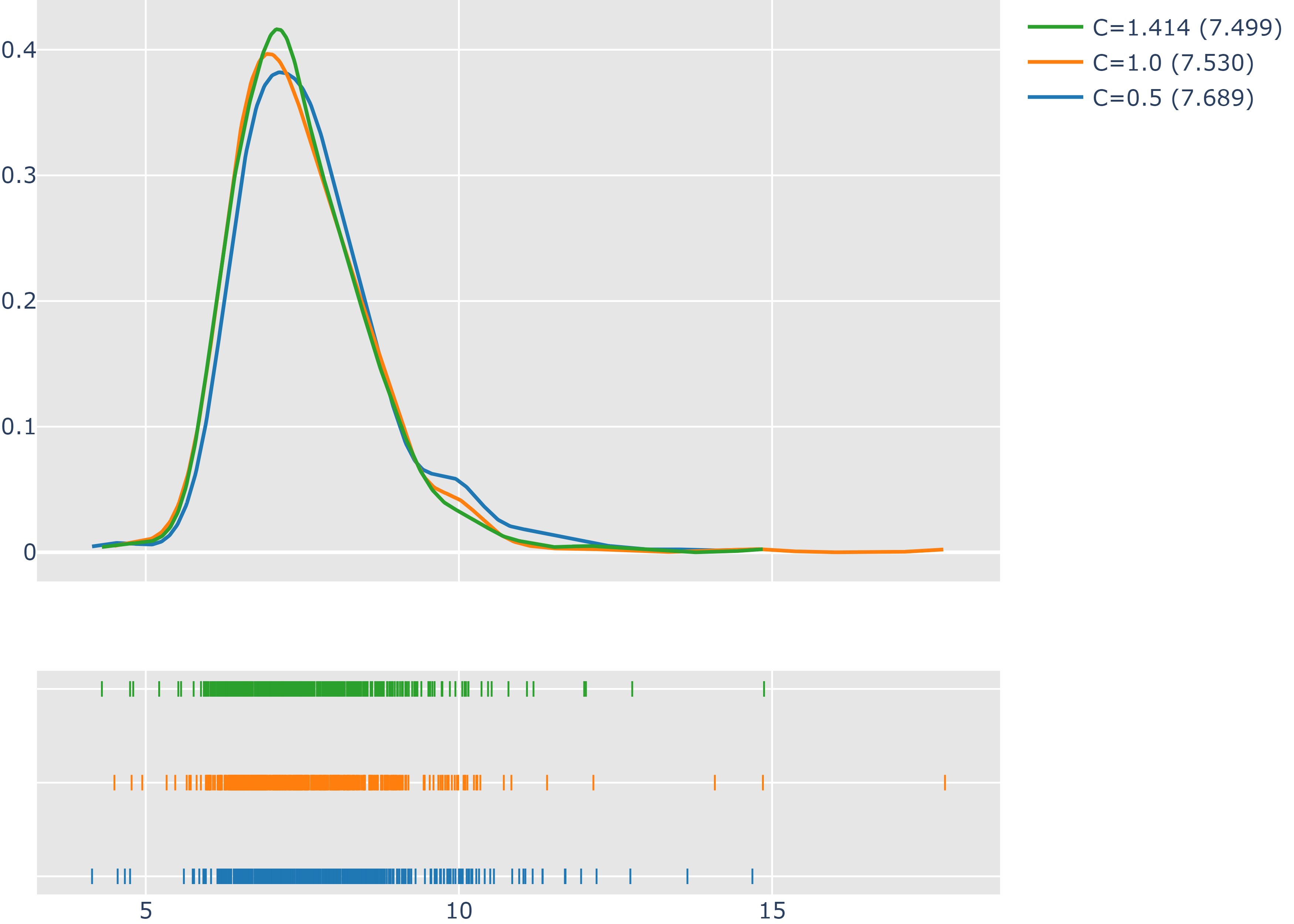}
    \caption{The distribution of the average depth of leaf nodes per example in MATH.}
    \label{fig:d_dist_math}
\end{figure*}

\begin{figure*}[t]
    \centering
    \includegraphics[width=\textwidth]{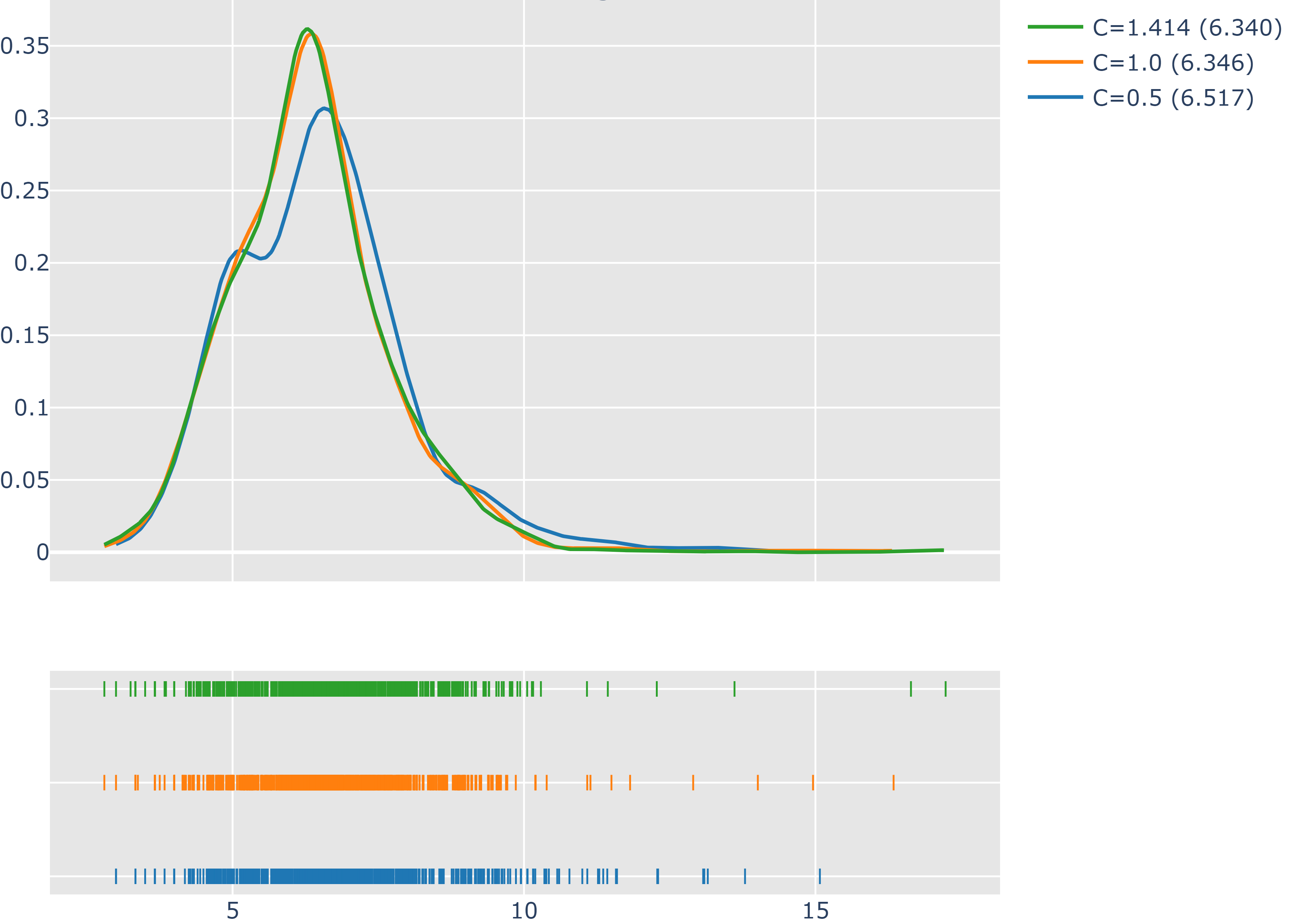}
    \caption{The distribution of the average depth of leaf nodes per example in SVAMP.}
    \label{fig:d_dist_svamp}
\end{figure*}

\begin{figure*}[t]
    \centering
    \includegraphics[width=\textwidth]{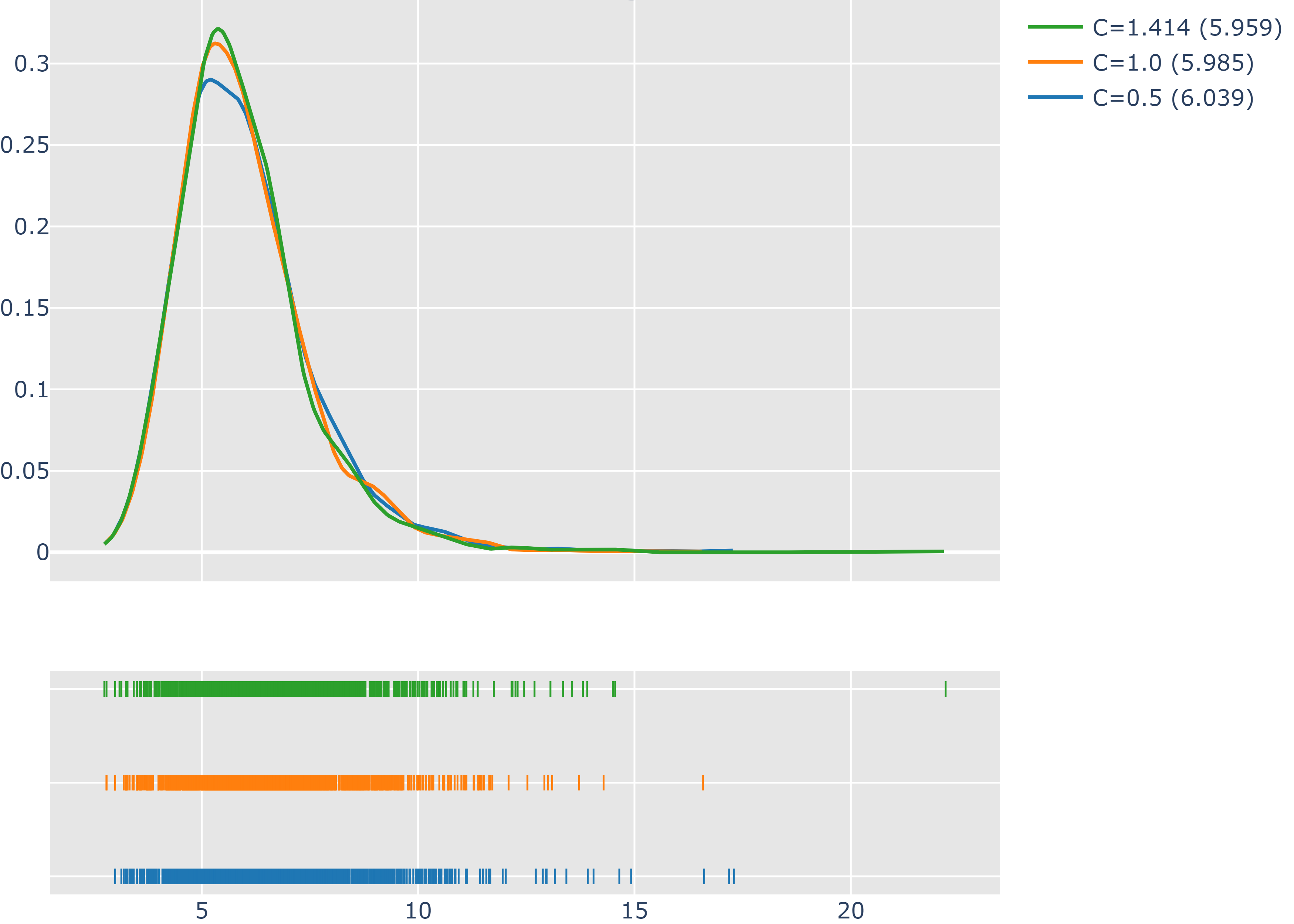}
    \caption{The distribution of the average depth of leaf nodes per example in ASDiv.}
    \label{fig:d_dist_asdiv}
\end{figure*}

\begin{figure*}[t]
    \centering
    \includegraphics[width=\textwidth]{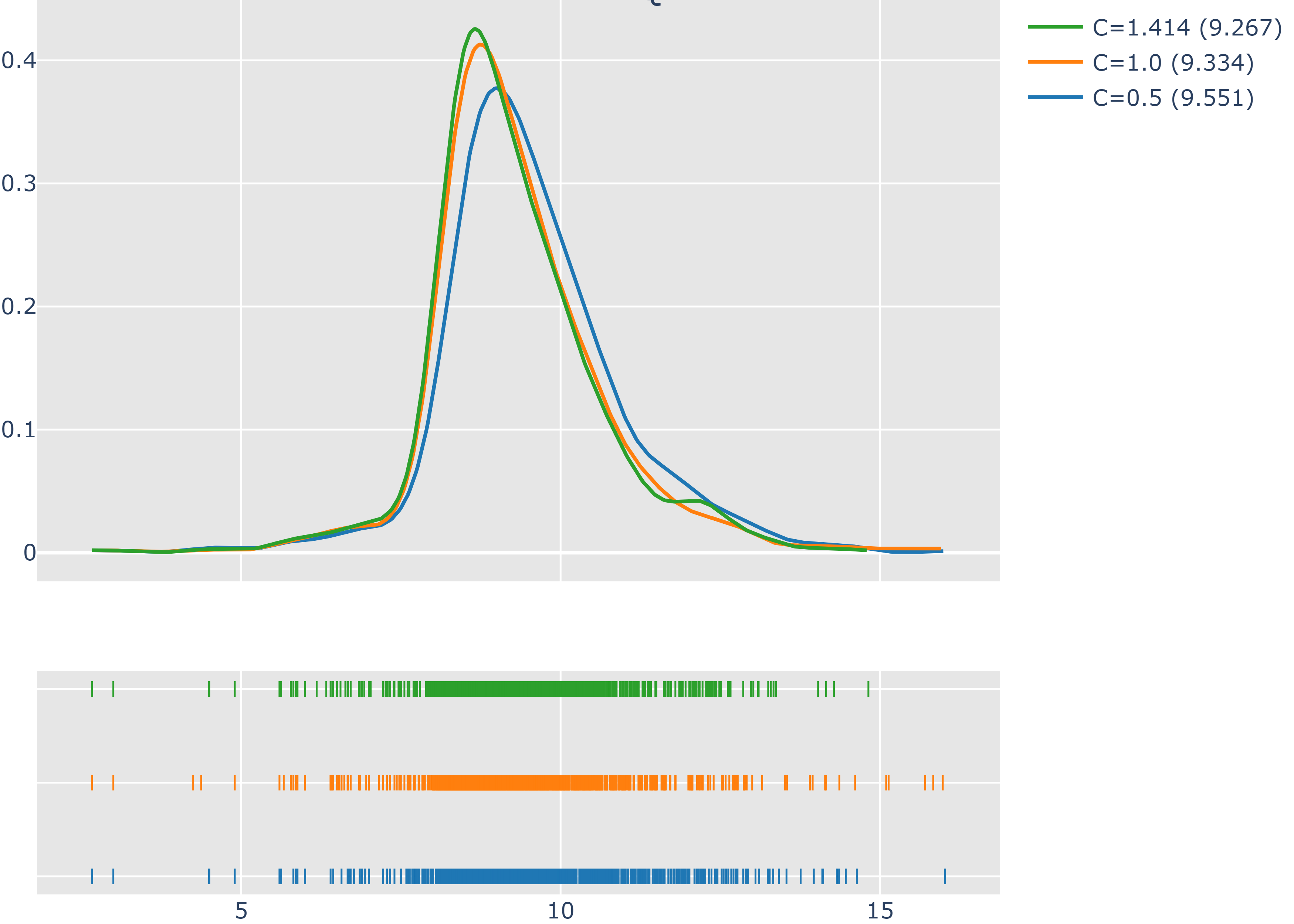}
    \caption{The distribution of the average depth of leaf nodes per example in MQA.}
    \label{fig:d_dist_mathqa}
\end{figure*}

Similar to Appendix~\ref{text:appx:r_dist}, we measured the average depth of leaf nodes for \( C = \{0.5, 1.0, 1.414\}\).
Specifically, we first measured the depth of all leaf nodes in the tree, then computed the average per sample and reported the results in the KDE plots Figs.~\ref{fig:d_dist_gsm8k}-\ref{fig:d_dist_mathqa}.
The value in parentheses represents the mean of all points in the plot. 
The results show that the length of reasoning trajectories increases as \( C \) decreases.
These results support our claim that LE-MCTS tends to perform more in-depth reasoning when a lower UCT constant \( C \) is used.
\begin{figure*}[t]
    \centering
    \includegraphics[width=\textwidth]{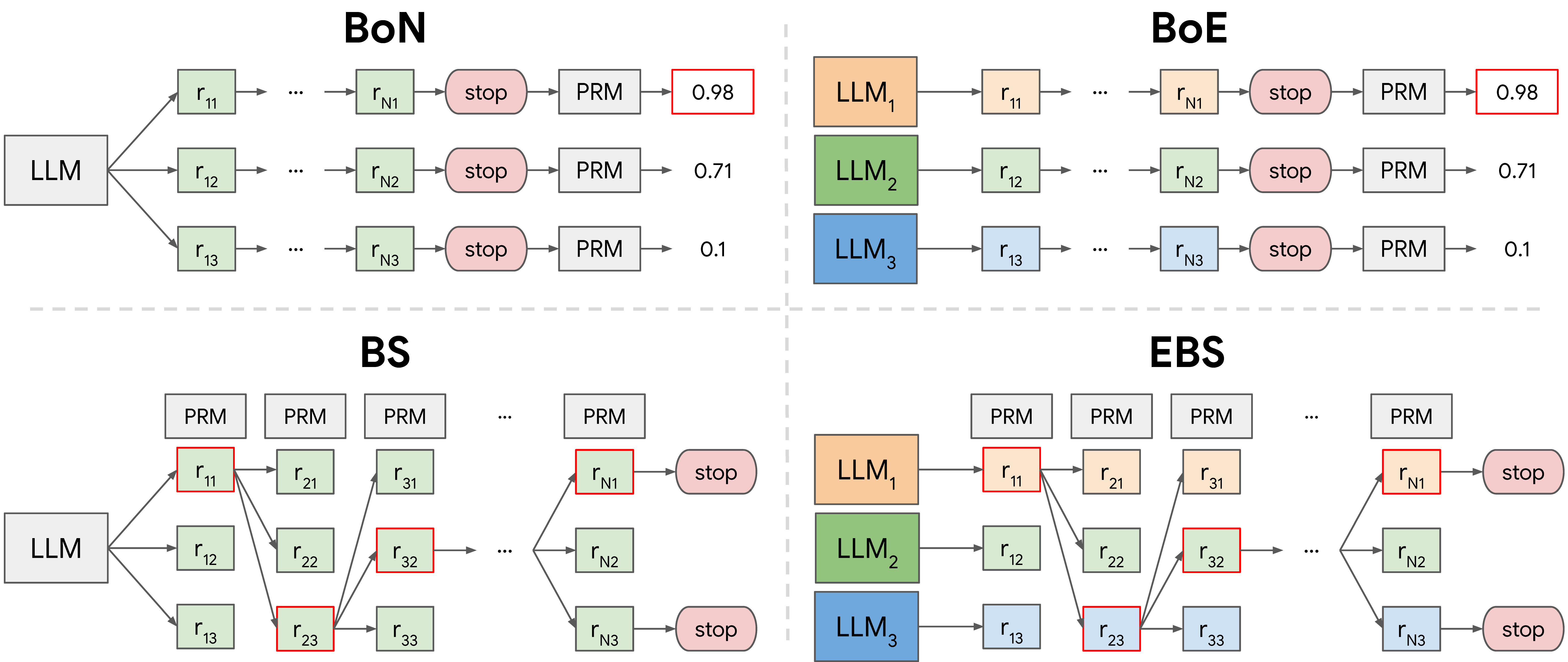}
    \caption{An illustration of process reward-guided decoding algorithms with \( N=3 \) and a beam size of 1.}
    \label{fig:prgbaselines}
\end{figure*}
\vspace{-1mm}
\section{Detailed Explanation of Baselines}
\label{text:appx:baselines}
In this section, we briefly describe the baseline methods to make our paper self-contained.
\medskip
\par \noindent \textbf{Process reward-guided decoding algorithms for a single LLM and ensemble language models} \;
Best-of-N (BoN)~\citep{stiennon-etal-2020-learning, lightman-etal-2023-prm800k} is a popular algorithm for aligning language models to human preferences. 
During inference, \( N \) samples are drawn from the language model, and the sample with the highest reward, determined by a reward model, is returned as the final output. 
BoN has also been applied to math problems~\citep{lightman-etal-2023-prm800k, yuan-etal-2023-scaling}, demonstrating its effectiveness.
Process reward-guided Beam Search (BS) was first proposed in OVM~\citep{ma-etal-2023-let}, which has employed a beam search strategy guided by OVM. 
Unlike conventional beam search, which relies on token-level probability, BS is steered by the estimated process reward at each step. 
We followed the BS implementation in OVM but used PRM from Math-Shepherd as the reward model, instead of OVM, to compute a consistent reward signal across all baselines.

Best-of-Ensemble (BoE) and Ensemble Beam Search (EBS) are simple modifications of BoN and BS, as depicted in Fig.~\ref{fig:prgbaselines}. 
For BoE, instead of using a single LLM, we employ multiple LLMs to generate candidates and select the best one among them. 
For EBS, at each step, multiple LLMs generate candidates for beam search. 
These straightforward modifications of existing process-reward guided decoding algorithms are significantly more effective than the original methods, as they provide a wider search space by incorporating different LLMs.
For all models, we set \( N=9 \) and sample candidates using nucleus sampling with a temperature 0.5.
\medskip
\par \noindent \textbf{LLM-Blender}~\cite{jiang-etal-2023-llm} \; 
LLM-Blender consists of two modules: PairRanker and GenFuser. 
PairRanker employs a specialized pairwise comparison method to distinguish subtle differences between candidate outputs. 
It jointly encodes the input text and a pair of candidates, using cross-attention encoders to determine the superior one. 
Then, GenFuser aims to merge the top-ranked candidates, generating an improved
output by capitalizing on their strengths and mitigating their weaknesses.
PairRanker employs DeBERTa~\citep{he-etal-2021-deberta} and GenFuser employs Flan-T5-XL~\citep{chung-etal-2022-flan} as the backbone and both are fine-tuned on the MixInstruct dataset proposed in LLM-Blender.
\medskip
\par \noindent \textbf{Mixture-of-Agents (MoA)}~\cite{wu-etal-2024-mixture} \; 
MoA employs multiple LLMs to iteratively enhance generation quality. 
It constructs a layered architecture in which each layer consists of multiple LLM agents. 
Specifically, LLMs in the first layer, denoted as agents \( A_{1,1}, \dots, A_{1,n} \), independently generate responses to a given prompt. 
These responses are then passed on to the next layer of agents \( A_{2,1}, \dots, A_{2,n} \), which can reuse models from the first layer, for further refinement.
This iterative refinement process is repeated over several cycles to produce a more robust and comprehensive response. 
In the final layer, a primary aggregator synthesizes all the outputs to obtain a final solution.
\medskip
\par \noindent \textbf{Ensemble LLMs via Vocabulary Alignment (EVA)}~\cite{xu-etal-2024-bridging} \; 
EVA proposed a novel vocabulary alignment method and fused LLMs at the token level. 
Their approach is based on the observation that, although various LLMs have distinct vocabularies, they often share a significant number of overlapping tokens. 
EVA leverages these tokens as bridges by first extracting embeddings of the overlapping tokens and learning a mapping matrix to project these embeddings into a shared space. 
Subsequently, by computing similarity scores between tokens in these vocabularies, EVA derives the semantic projection matrix \( W \). 
This enables the projection of output distributions from one LLM to another, allowing for the generation of reasonable tokens based on the fused distribution of these LLMs at each inference step. 
Finally, they further enhance the approach by devising a filtering strategy to exclude models that generate unfaithful tokens.
\begin{table*}[ht]
\centering
\resizebox{\textwidth}{!}{
\begin{tabularx}{\textwidth}{X}
\hline
\textbf{Question}: Josh decides to try flipping a house. He buys a house for \$80,000 and then puts in \$50,000 in repairs. This increased the value of the house by 150\%. How much profit did he make? \\
\hline
\textbf{LE-MCTS, Correct answer} (\textit{reward}: 0.932) \\
\rule{0pt}{3ex}
\textit{LLaMA-3}: Let's break it down step by step! \\
\textit{Gemma-2}: 1. **Calculate the total investment:** Josh spent \$80,000 (house) + \$50,000 (repairs) = \$130,000 \\
\textit{Gemma-2}: 2. **Calculate the increase in value:** A 150\% increase means the house's value went up by 1.5 times its original price. So, the increase is \$80,000 * 1.5 = \$120,000 \\
\textit{Gemma-2}: 3. **Calculate the new value:** The house is now worth \$80,000 (original) + \$120,000 (increase) = \$200,000 \\
\textit{Rho-Math}: 4. **Calculate the profit:** The profit is the difference between the new value and the total investment. So, \$200,000 - \$130,000 = \$70,000 \\
\textit{LLaMA-3}: 5. **Answer:** Josh made a profit of \$70,000. \\
\hline
\textbf{LE-MCTS, Wrong answer} (\textit{reward}: 0.755) \\
\rule{0pt}{3ex}
\textit{Gemma-2}: Here's how to solve the house flipping problem: \\ 
\textit{Gemma-2}: **1. Calculate the total investment:** \\
\textit{Gemma-2}: * Josh spent \$80,000 on the house + \$50,000 on repairs = \$130,000 total investment. \\
\textit{LLaMA-3}: **2. Calculate the increased value of the house:** \\
\textit{Rho-Math}: \textcolor{red}{* 150\% of \$130,000 = 1.5 * 130,000 = 195,000. So the house is now worth 195,000.} \\
\textit{Gemma-2}: **3. Calculate the profit:** \\
\textit{Gemma-2}: *  The house is now worth \$195,000 - \$130,000 (investment) = \$65,000 profit. \\
\textit{LLaMA-3}: The answer is \$65,000. \\
\hline
\textbf{BoE, Wrong answer} (\textit{reward}: highest among candidates)\\
\rule{0pt}{3ex}
The original value of the house was \$80,000. He put in an additional \$50,000 in repairs. This increased the value of the house by 150\%. \textcolor{red}{So the new value of the house is \$80,000 + 150\% of \$80,000 + \$50,000. This is \$80,000 + 1.5 * \$80,000 + \$50,000.} This is \$80,000 + \$120,000 + \$50,000. This is \$250,000.  So he made a profit of \$250,000 - \$130,000 = \$120,000. \\
\hline
\end{tabularx}
}
\caption{An example of the outputs generated by LE-MCTS and BoE to solve a test case in GSM8K. We highlight the first error made by the model in red.}
\label{tab:case_study}
\end{table*}
\section{Tasks and Dataset Statistics}
\label{text:appx:dataset}
\medskip
\par \noindent \textbf{GSM8K}~\citep{cobbe-etal-2021-gsm8k} \; Grade School Math 8K (GSM8K) is a dataset of 8.5K high-quality linguistically diverse grade school math word problems. 
These problems take between 2 and 8 steps to solve. 
Solutions primarily involve performing a sequence of elementary calculations using basic arithmetic operations to reach the final answer. 
We used 1,319 test examples for the experiment.
\medskip
\par \noindent \textbf{MATH}~\citep{hendrycks-etal-2021-math} \; MATH is a dataset of 12.5k challenging competition mathematics problems. 
Each problem in MATH has a full step-by-step solution and requires a wide range of mathematical problem solving abilities at varying levels of difficulty and topics. 
Although the original MATH dataset provides 5,000 test examples, PRM800K~\citep{lightman-etal-2023-prm800k} used 4,500 of these for training. 
Following standard practice~\citep{lightman-etal-2023-prm800k,wang-etal-2024-math}, we used the remaining 500 examples (MATH500) for evaluation to prevent data leakage.
\medskip
\par \noindent \textbf{SVAMP}~\citep{patel-etal-2021-nlp} \; Simple Variations
on Arithmetic Math word Problems (SVAMP) is a dataset of 1,000 arithmetic word problems with grade level up to 4 by applying simple variations over word problems in an existing dataset. 
The authors modified the ASDiv-A~\citep{miao-etal-2020-diverse} dataset with three variations: changing the question object and structure, tweaking the underlying reasoning, and shuffling objects and phrases.
We used the entire dataset for the evaluation.
\medskip
\par \noindent \textbf{ASDiv}~\citep{miao-etal-2020-diverse} \; Academia Sinica Diverse MWP Dataset (ASDiv) is a dataset of 2,305 math word problems commonly taught in elementary school.
ASDiv focuses on the diversity of math problems in terms of text patterns and problem types.
We used the entire dataset for the evaluation.
\medskip
\par \noindent \textbf{MQA}~\citep{amini-etal-2019-mqa} \; MathQA (MQA) is a dataset of 37k English multiple choice math word problems that cover multiple math domain categories by modeling operation programs corresponding to word problems in the AQuA dataset~\citep{ling-etal-2017-program}. 
Specifically, MQA manually annotated GRE- and GMAT-level math word problems in AQuA with formal operation programs.
We used the 1,000 examples provided by the math-evaluation-harness library for the evaluation.

\section{Case Study}
\label{text:appx:case_study}
We present an example generated by LE-MCTS and BoE using a test case from GSM8K in Table~\ref{tab:case_study}.
BoE, the second-best ensemble model on GSM8K, made an error in the intermediate reasoning step, "\textit{So the new ... \$50,000.}".
A similar mistake appears in the trajectory discovered by LE-MCTS, despite receiving a high reward of 0.755.
However, the highest-reward reasoning chain in LE-MCTS (\textit{i.e.}, the path that reaches the leaf node with a reward of 0.932) demonstrates perfect reasoning.
The provided example illustrates the flexibility of LE-MCTS in reasoning. 
Although LE-MCTS can make mistakes during the search, it is capable of identifying better reasoning paths by incorporating a lookahead search and diverse reasoning steps from multiple LLMs.


\end{document}